%% file: main.tex
\documentclass[10pt,twocolumn,letterpaper]{article}
\usepackage{mathtools, nccmath}

\usepackage[pagenumbers]{cvpr} 


\definecolor{cvprblue}{rgb}{0.21,0.49,0.74}
\usepackage[pagebackref,breaklinks,colorlinks,allcolors=cvprblue]{hyperref}

\def\paperID{11846} 
\def\confName{CVPR}
\def\confYear{2025}

\graphicspath{ {./images/} }
\DeclarePairedDelimiter{\nint}\lfloor\rceil
\usepackage{tabularx}
\usepackage{algorithm}
\usepackage{algpseudocode}
\usepackage{nicefrac}
\usepackage{multirow}
\usepackage{caption} 
\title{Data-Free Group-Wise Fully Quantized Winograd Convolution via Learnable Scales}

\author{Shuokai Pan \qquad Gerti Tuzi \qquad Sudarshan Sreeram \qquad Dibakar Gope\\
Arm Inc.\\
{\tt\small \{shuokai.pan, dibakar.gope\}@arm.com}
}

\begin{document}
\maketitle
\input{sec/abstract}
\input{sec/intro}

\input{sec/related_work}
\input{sec/winograd_convolution}
\input{sec/learnable_scales}

\input{sec/evaluation}
\input{sec/conclusion}


{
    \small
    \bibliographystyle{ieeenat_fullname}
    \bibliography{main}
}

\input{supplementary_arxiv}

\end{document}

%% file: sec/abstract.tex
\begin{abstract}
Despite the revolutionary breakthroughs of large-scale
text-to-image diffusion models for complex vision and downstream tasks, their extremely high computational and storage costs limit their usability. Quantization of diffusion models has been explored in recent works to reduce compute costs and memory bandwidth usage. To further improve inference time, fast convolution algorithms such as Winograd can be used for convolution layers, which account for a significant portion of computations in diffusion models. However, the significant quality loss of fully quantized Winograd using existing coarser-grained post-training quantization methods, combined with the complexity and cost of finetuning the Winograd transformation matrices for such large models to recover quality, makes them unsuitable for large-scale foundation models. Motivated by the presence of a large range of values in them, we investigate the impact of finer-grained group-wise quantization in quantizing diffusion models. While group-wise quantization can largely handle the fully quantized Winograd convolution, it struggles to deal with the large distribution imbalance in a sizable portion of the Winograd domain computation. To reduce range differences in the Winograd domain, we propose finetuning only the scale parameters of the Winograd transform matrices without using any domain-specific training data. 
Because our method does not depend on any training data, the generalization performance of quantized diffusion models is safely guaranteed. 
For text-to-image generation task, the $8$-bit fully-quantized diffusion model with Winograd provides near-lossless quality (FID and CLIP scores) in comparison to the full-precision model. This, coupled with the development of highly optimized kernels for group-wise fully quantized Winograd, improves CPU wall-clock time by $31.3\%$ when compared to the convolution layers of a diffusion model.
For image classification, our method outperforms the state-of-the-art Winograd PTQ method by $1.62\%$ and $2.56\%$ in top-1 ImageNet accuracy on ResNet-18 and ResNet-34, respectively, with Winograd F(6, 3). 
\end{abstract}

%% file: sec/intro.tex
\section{Introduction}
\label{sec:intro}
In recent years, foundational pre-trained diffusion models have gained a rapid rise in popularity in the field of image generation due to their ability to produce complex and incredibly detailed high-quality photorealistic images from natural language prompts~\cite{song2022denoisingdiffusionimplicitmodels,chitwan2022imagen,rombach2022highresolutionimagesynthesislatent,podell2024sdxl}. Furthermore, foundation DMs have successfully demonstrated their flexibility in supporting and achieving high-quality performance on a wide range of downstream computer vision tasks, such as image edition, style transformation, image super-resolution, image-to-image translation, and many others, in contrast to previous Generative Adversarial Networks-based image generation models that suffered from unstable training~\cite{prafulla2021gans}. However, diffusion models typically require many denoising steps and forward passes to convert Gaussian noises into real images using neural network layers with over $1$ billion parameters.
Therefore, deploying these large-scale diffusion models to on-device for inference has been a significant challenge due to their unprecedented size, memory bandwidth, and compute cost requirements~\cite{sui2024bitsfusion}.


\begin{figure*}[h!]
    \centering
    \includegraphics[width=0.75\linewidth]
    {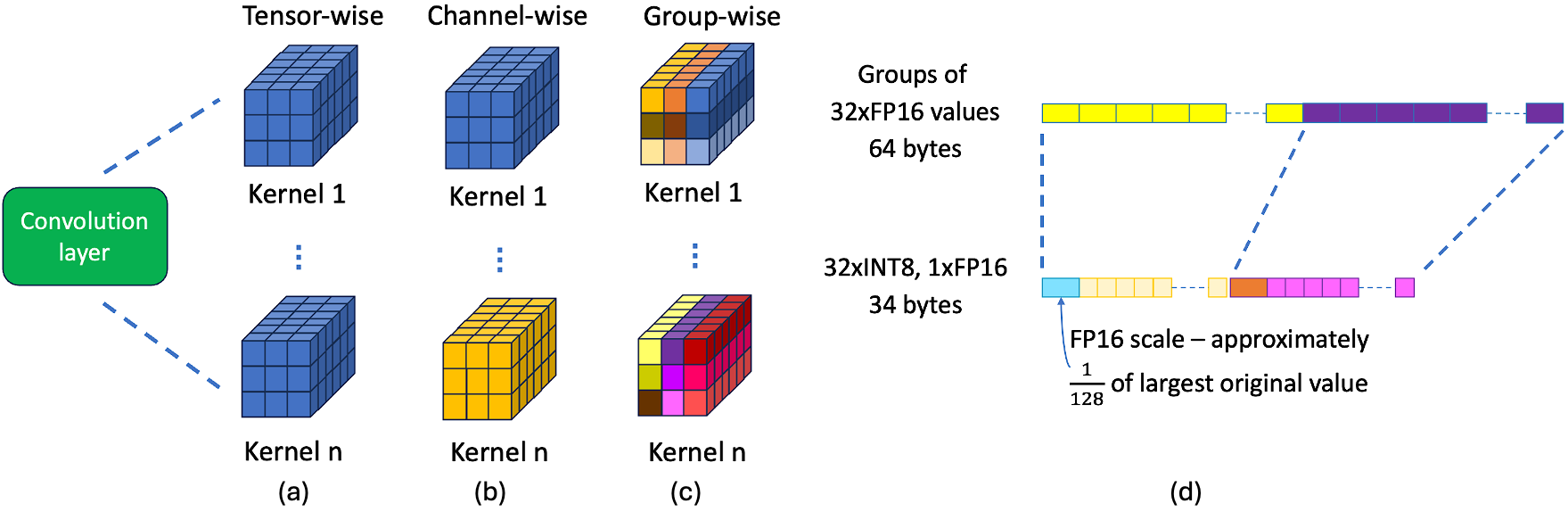}
    \caption{Group-wise quantization for convolution layers.}
    \label{fig:group-quant}
\end{figure*}

Quantization has proven to be an effective method for converting high-precision ($16$ or $32$-bit) model weights and activations to lower-precision values, such as 8-bit integers, reducing the model's memory and computational requirements while maintaining accuracy, making it more suitable for deployment on devices with limited resources. 
Among different quantization methods, data-free post-training quantization (PTQ) compresses model parameters after training. While previous studies attempted to quantize weights and activations in diffusion models using coarse-grained PTQ techniques~\cite{shang2023posttrainingquantizationdiffusionmodels, Li_2023_ICCV, he2023ptqdaccurateposttrainingquantization, wang2024accurateposttrainingquantizationdiffusion, chen2024qditaccurateposttrainingquantization, wang2024quest, tang2024ptq, zhao2024mixdq}, such as tensor-wise or channel-wise quantization, they often resulted in tangible loss of quality, especially under low-bit settings. One issue with these coarser-grained quantization approaches is that outlier values, i.e., outliers, can have a disproportionate impact on scaling: the full range of the lower-precision data type is not used effectively, which lowers the quantized model’s accuracy. Finer-grained group-wise quantization instead has shown great promise in successfully quantizing and compressing large-scale generative AI foundation models, particularly large language models. Group-wise quantization~\cite{dai2021vsquant} has a finer granularity than standard tensor-wise or channel-wise quantization, allowing it to reduce quantization noise natively while approaching the high-precision (floating point) quality of a foundation model. Group-wise quantization quantizes in groups, whereby weights are divided into groups of $32$, $64$, or $256$. Each group is then quantized individually to mitigate the effect of outliers and increase precision.

In this work, in an attempt to preserve image generation quality, we investigate the impact of group-wise quantization of weights and activations in large-scale diffusion models while ensuring high runtime performance through the development of highly optimized matrix multiply kernels for group-wise quantized diffusion models.

While group-wise quantization accelerates inference, the high computational costs of these large-scale foundation models necessitate further speedups to meet response time requirements and pave the way for the deployment of them on edge or mobile devices. Convolution operations account for a significant fraction of the computation time in large-scale diffusion models during inference and training. Different algorithmic methods have been devised to speed up this core operation. Among other compression techniques for reducing this cost, 
fast convolution algorithms in replacement of direct convolutions, such as Winograd convolution algorithms~\cite{lavin2015fastalgorithmsconvolutionalneural}, are observed to accelerate the widely used small-size convolution and can provide additional speedups. 
While Winograd convolution computation in the quantized domain can significantly accelerate diffusion models, its use in the quantized context results in a significant increase in quantization noise and a subsequent drop in quality, as observed in previous studies~\cite{meng2019complexwinograd, fernandez2020searching, bqw, chen2024towards}. The use of group-wise quantization can largely resolve the quantization and associated numerical error problem of input transformation and element-wise multiplications in the Winograd domain. However, it cannot quantize the intermediate values for output transformation, which account for a significant portion of the compute time, in the Winograd domain well due to the large range differences in values, which can result in a significant degradation in quality when applying group-wise quantization to Winograd convolutions.
Previous studies either used costly learning of the Winograd transformation matrices to recover network accuracy~\cite{fernandez2020searching} or performed quantization-aware training or finetuning of the Winograd matrices with a domain-specific dataset~\cite{chen2024towards}, which could lead to overfitting. They may not perform well on previously unseen datasets, particularly for foundation models and various downstream tasks in the production setting. QAT can cause overfitting to small training data sets, so it is generally not recommended for foundational models to maintain their pristine quality.

In this work, we show that the small set of scale factors associated with Vandermonde matrices that derive the Winograd transformation matrices can only be altered to limit the large distribution imbalance of intermediate values in the Winograd domain. Furthermore, these scale factors can be finetuned using random noise without the need for any domain-specific dataset. We derive this requirement from a theoretically grounded analysis. The generalization ability of our data-free group-wise quantized Winograd is inherently guaranteed since it does not need any input data.





The key contributions of this work are as follows:
\begin{itemize}
\item We propose a novel hardware-aware quantized Winograd convolution algorithm that uses group-wise quantization to fully quantize all Winograd pipeline operations. To the best of our knowledge, this is the first empirical demonstration of quantized Winograd convolution's impact on large-scale image generation models.
\item We propose fine-tuning only the scale parameters of the Winograd transformation to reduce dynamic range differences.
Data-free finetuning makes things transferable to other datasets, which is a prerequisite for foundational models. To the best of our knowledge, we are the first to perform full quantization of Winograd convolution with large tile sizes and ensure high quality without using any training or calibration data.
\item We restrict the group size to be either an integer multiple or at least equal to a processor's vector width to benefit from vectorization. Our group-wise quantized matrix multiply kernels for Winograd convolution, developed as part of this work, can take advantage of the efficient vectorized 
matrix multiply operations in a processor, significantly improving the runtime performance of convolution and transformer layer operations.
\item Our group-wise $8$-bit fully quantized diffusion models with Winograd convolutions in conjunction with learned scales provide comparable image generation performance (very small FID change).
It also demonstrates superior classification accuracy ($2.56\%$ improvement in accuracy for the ImageNet dataset on ResNet34) when compared to state-of-the-art approaches, while optimized kernels ensure a $31.3\%$ improvement in runtime performance over 
standard convolutions ($12.8\%$ improvement for end-to-end diffusion models). 
\end{itemize}

%% file: sec/related_work.tex
\section{Related work}
\label{sec:related_work}

\begin{figure*}[t]
\centering
\includegraphics[width=0.65\paperwidth]
{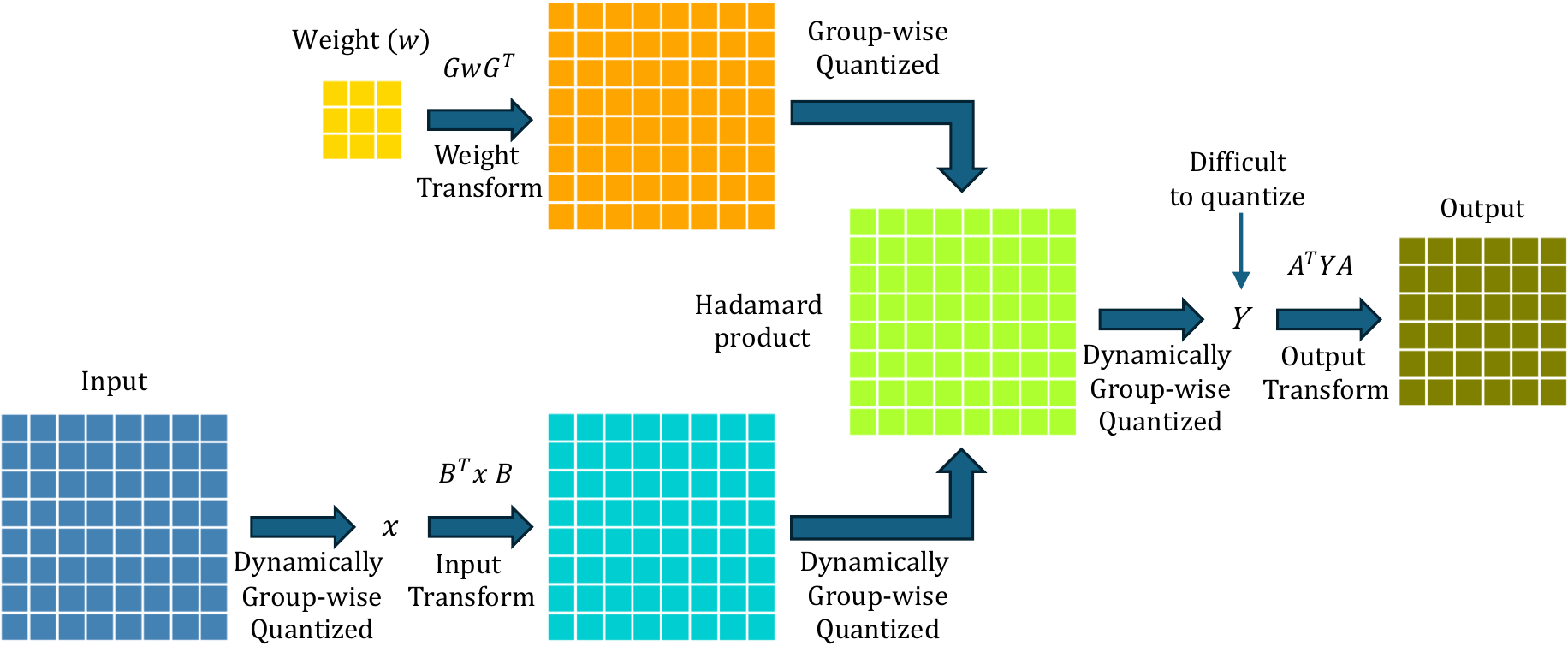}
\caption{Group-wise fully quantized Winograd convolution. 
Applying group-wise quantization to Hadamard 
product and input transformation has a minimal impact on model quality. However, doing the same with the output transformation leads to a significant drop in model accuracy. Weight transformation can be done offline with high precision.}
\label{fig:winograd_conv}
\end{figure*}

\textbf{Diffusion models. }
Diffusion models~\cite{chitwan2022imagen}
can produce high-quality images using an iterative denoising process. They primarily consist of a text encoder, a denoising neural network, such as UNet, in which the convolution layers can take up a significant portion of the time, and an image decoder.
Although denoising diffusion models have demonstrated phenomenal capabilities in a range of generative tasks, their slow generation speed prevents their widespread deployment. 
This can be attributed to their lengthy iterative denoising process through the UNet and high computational demand of the UNet at each step.
While numerous studies have been conducted to accelerate this sampling process~\cite{luhman2021knowledgedistillationiterativegenerative, salimans2022progressivedistillationfastsampling, yin2024onestepdiffusiondistributionmatching, liu2024instaflowstephighqualitydiffusionbased} and design fast samplers~\cite{song2022denoisingdiffusionimplicitmodels, song2021scorebasedgenerativemodelingstochastic, lu2022dpmsolverfastodesolver}, in this work we investigate model quantization on diffusion models, which is an orthogonal direction to the above methods and can significantly reduce the computational complexity of the denoising network at each sampling step, thereby further accelerating the sampling process.

\textbf{Diffusion model quantization. }
Model quantization can be categorized into two types: quantization-aware training (QAT) and post-training quantization (PTQ). QAT 
often has negligible impacts on model quality after quantization. However, this requires the original training pipeline and datasets, which can be very challenging to set up, especially for large-scale generative AI foundation models. PTQ, on the other hand, applies model quantization after training and usually only requires a small number of samples for calibration of the quantization parameters. Hence, it is much less time-consuming and computationally intensive, generally more favored in network deployment, and is the focus of this work. However, PTQ can lead to significant model quality degradation if not applied carefully. Many PTQ methods have thus been proposed to tackle this issue, for instance, by minimizing the 
reconstruction errors of tensors before and after quantization \cite{nagel2020downadaptiveroundingposttraining, li2021brecqpushinglimitposttraining}. 
The uniform quantization of a floating-point tensor $x$ into $b$-bit integer can be written as, 
\begin{align}
    x_{int} &= clamp( \nint{\frac{x}{s}}, c_{min}, c_{max} ) \nonumber \\
    x &\approx s * x_{int} = \hat{x}    
\end{align}
PTQ of diffusion models has been studied in a number of previous works~\cite{shang2023posttrainingquantizationdiffusionmodels, Li_2023_ICCV, he2023ptqdaccurateposttrainingquantization, wang2024accurateposttrainingquantizationdiffusion, chen2024qditaccurateposttrainingquantization, wang2024quest, tang2024ptq, zhao2024mixdq, so2023temporal,tangeccv2024pcr}. One of the key observations is that the activation distributions of the noise estimation network change greatly over the sampling time steps~\cite{Li_2023_ICCV}. \cite{wang2024quest} emphasizes the importance of activation outliers. Thus, advanced calibration processes have been proposed in these works, but they can be quite complicated and are not easily transferable from one architecture to another~\cite{chen2024qditaccurateposttrainingquantization}. In this work, we proposed to tackle this problem with the more flexible group-wise quantization, which is inherently more robust to distribution changes because the activation quantization parameters are computed for each group of values and dynamically. 

\textbf{Quantized Winograd convolution. }Winograd algorithms~\cite{lavin2015fastalgorithmsconvolutionalneural} are fast convolution algorithms based on minimal filtering theory \cite{winograd1980arithmetic}, and they are well-known for being the fastest implementation of small convolution kernels found in modern neural networks. The Winograd algorithm, similar to the FFT, converts tiles of input activations and weight elements into the Winograd domain before performing element-wise multiplications or Hadamard product, which reduces theoretical computation complexity, as shown in Figure~\ref{fig:winograd_conv}. In general, the larger the tile size, the greater the reduction in computational complexity. However, this is not always preferred because larger tile sizes result in greater numerical errors due to the exponentially increasing values of the Winograd transformation matrices as tile size increases. This is also the main reason why Winograd algorithms are typically implemented using $32$-bit floating-point arithmetic and with relatively small tile sizes, such as $4\times4$.

Previous studies~\cite{meng2019complexwinograd, fernandez2020searching, bqw, chen2024towards, li2021lowino} have investigated combining Winograd convolution with quantization; however, some \cite{li2021lowino} only quantized the Hadamard multiplication while doing the input and output transformations in floating-point arithmetic. \cite{fernandez2020searching} proposes a flexible Winograd convolution scheme by treating the transformation matrices as learnable parameters. However, this requires a full training pipeline setup, similar to the complexity of QAT. Instead of expensive QAT, \cite{chen2024towards} performs a two-stage finetuning process to finetune the Winograd transformation matrices with a small set of training samples. However, it cannot fully restore full-precision model accuracy for larger Winograd $F(6, 3)$. Our work nearly bridges the accuracy gap with the full-precision model while fully quantizing Winograd convolution without requiring any training samples or domain-specific data.

%% file: sec/winograd_convolution.tex
\renewcommand{\algorithmicrequire}{\textbf{Input:}}
\renewcommand{\algorithmicensure}{\textbf{Output:}}

\section{Group-wise fully quantized Winograd convolution}
\label{sec:winograd_convolution}

\subsection{Winograd convolution}
Using Winograd, the computation of a 2D convolution output tile, $y$, of size $m \times m$ with a kernel filter of size $r \times r$, as presented in \cite{lavin2015fastalgorithmsconvolutionalneural} as $F(m \times m, r \times r)$, is often given by,
\begin{equation}
    y = A^T \Bigl[ [GwG^T] \odot [B^TxB] \Bigr] A
\end{equation}
or in more details, 
\begin{equation}
    W = GwG^T
\label{weight_transform}
\end{equation}
\begin{equation}
    X = B^TxB
\label{input_transform}
\end{equation}
\begin{equation}
    Y = W \odot X
\label{Hadamard}
\end{equation}
\begin{equation}
    y = A^T Y A
\label{output_transform}
\end{equation}

where $w$ is a $r \times r$ filter, and $x$ is an $(m+r-1) \times (m+r-1)$ input tile. $B$, $G$, and $A$ are called Winograd transformation matrices, where $B$ and $G$ transform the weights and input feature maps, respectively, from the spatial domain to the Winograd domain, and $A$ transforms the output feature maps ($Y$) back to the spatial domain, after the element-wise multiplication denoted by $\odot$. This is also often short-handed to $F(m, r)$ when the tile and filter are square matrices, which is used in this work. 

The Winograd transformation matrices can be constructed from the Chinese remainder theorem by choosing $n=m+r-1$ pairs of so-called polynomial points or Lagrange interpolation points $(f_i, g_i)$. The matrices can then be derived from their Vandermonde matrix $V$, as shown below. 

\begin{equation}
A^T = V_{n \times m}^T S_A
\label{scale_a}
\end{equation}
\begin{equation}
B^T = S_B V_{n \times n}^{-T}
\label{scale_b}
\end{equation}
\begin{equation}
G = S_G V_{n \times r}
\label{scale_g}
\end{equation} 

where $S_A$, $S_B$, and $S_G$ are diagonal square matrices satisfying the following condition. For more details of derivation, please refer to \cite{vincent2017improving}. 
\begin{equation}
S_A S_B S_G = I 
\label{scaling_condition}
\end{equation}

\begin{figure*}[t]
    \centering
    \includegraphics[width=0.975\linewidth]
    {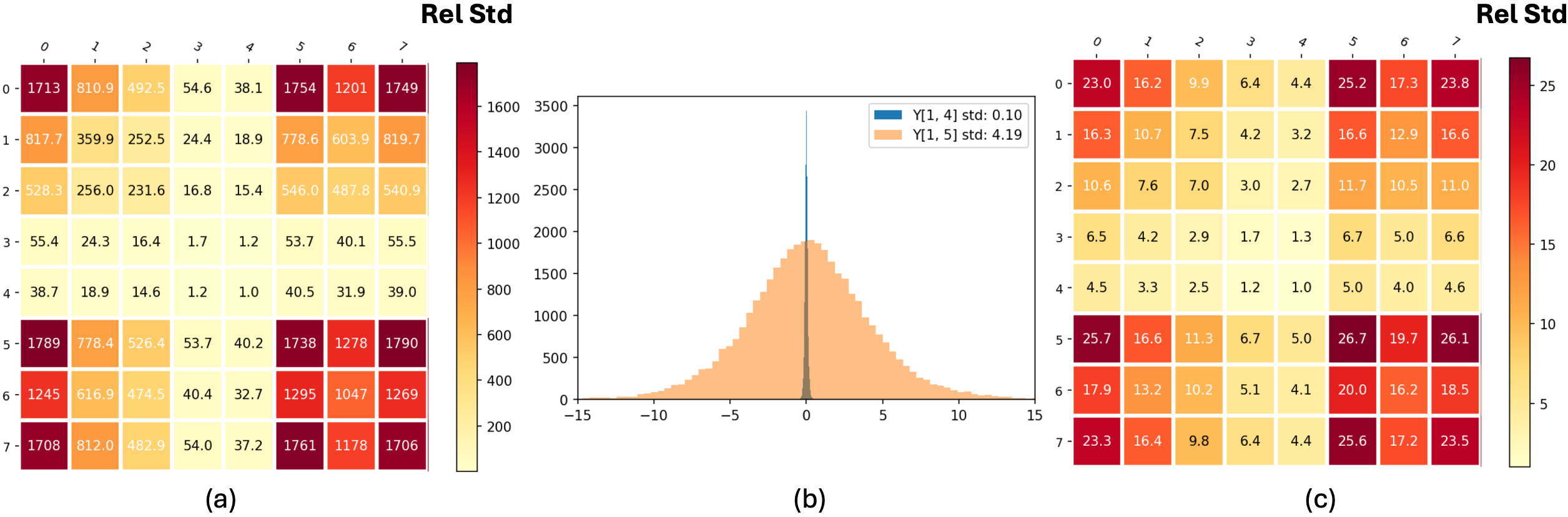}
    \caption{Dynamic ranges across different taps or pixels of the Winograd domain output (Y) are very different. (a) Relative standard deviations at all of the $8 \times 8$ locations of the Winograd domain output tile, obtained from the InstaFlow-0.9B model. (b) Histograms of values at locations (1, 4) and (1, 5). (c) Relative standard deviations after learning Winograd scales.}
    \label{fig:winograd_dist}
\end{figure*}
\subsection{Group-wise quantization and optimized kernels}
Naive application of model quantization to Winograd convolutions is often a challenging task, as shown in other studies \cite{fernandez2020searching, chen2024towards}. This is mainly because the quantization errors would exacerbate the numerical errors of Winograd convolution, especially for large tile sizes, thus usually leading to significant model accuracy degradations. 

Inspired by the recent advances in LLM quantization, we adopt a fine-grained group-wise quantization scheme for both the weights and activations because of its reduced quantization errors and minimal dependence on the calibration data. 
Tensors are first divided into hardware vectorization-friendly groups, as illustrated in Figure~\ref{fig:group-quant}, and each group is then quantized individually to reduce the impact of outliers and improve precision.
We also apply this for the Winograd input transform (Eq.~\ref{input_transform}) and output transform (Eq.~\ref{output_transform}), resulting in a fully quantized Winograd convolution pipeline. 

In addition, we develop a set of highly optimized matrix multiply kernels for group-wise quantized $8$-bit diffusion models to unleash the full potential of modern CPUs. These kernels can fully take advantage of available vector and matrix multiply instructions to maximize MAC unit utilization, amortize the cost of loading the operands, minimize overhead and memory accesses, and achieve the best possible performance (to date) on CPUs for diffusion models. Our highly optimized group-wise quantized kernels can achieve similar runtime performance as that of the coarser-grained quantization methods while offering better quality.

%% file: sec/learnable_scales.tex
\section{Learnable scales for group-wise qantized Winograd convolution}
\label{sec:learnable_scales}
\subsection{Learnable Winograd transform scales}
Because of the fine granularity of group-wise quantization, we observe that direct application of it to the Winograd input transform and Hadamard product computation generally would not have a significant impact on model quality. 
However, this is not the case for the output transform.
This is mainly because of the huge dynamic range differences across different taps or pixels of the Winograd domain output, $Y$ (Eq.~\ref{Hadamard}), as shown in Figure~\ref{fig:winograd_dist}(a). In order to utilize efficient integer arithmetic 
operations,
there should be either a single scale factor for the entire output tile or one for each row or column. However, both would lead to significant quantization errors due to the `cross'-like dynamic range differences at the $8 \times 8$ locations. Although pixel-wise quantization can effectively reduce this quantization error, it precludes the efficient use of integer computation kernels.

Given the fact that the output feature map Y in the Winograd domain depends on inputs, original pre-trained weights, and the input transformation (B) and weight transformation (G) matrices, the large dynamic range of Y across pixels is primarily attributed to the values of G and W, as well as the variances of values in weights and inputs. In the absence of finetuning original weights in the PTQ setup, the large range differences may be effectively reduced by manipulating the two transformation matrices, B and G, and in turn their norm of rows.
From Eq.~\ref{scale_b} and Eq.~\ref{scale_g}, each row of the Vandermonde matrices of $B^T$ and $G$ are scaled by the diagonal scaling matrices $S_B$ and $S_G$, respectively, which are directly controlling the norms of the row vectors in $B^T$ and $G$. 

Following this intuition, we propose to reduce the quantization noise of Winograd output transform by learning the diagonal scaling matrices $S_B$ and $S_G$, while $S_A = (S_B S_G)^{-1}$, as given by Eq.~\ref{scaling_condition}, and can be easily computed. More formally, if we define the Vandermonde matrices as $V_B = V_{n \times n}^{-T}$, $V_G = V_{n \times r}$ and $V_A = V_{n \times m}^T$ to simplify notation, then the Winograd transformations can be rewritten as: 
\begin{equation}
    W = S_G V_G w V^T_G S_G
\end{equation}
\begin{equation}
    X = S_B V_B x V^T_B S_B
\end{equation}
\begin{equation}
    y = V_A S_A Y S_A V^T_A
\end{equation}

Then we apply group-wise quantization and integer matrix multiplication to all stages of Winograd convolution, specifically: 
\begin{align}
    X &\approx s_{qB} s_{qB} s_{qx} Q\Bigl(\frac{S_B V_B}{s_{qB}}\Bigr) Q\Bigl(\frac{x}{s_{qx}}\Bigr) Q\Bigl(\frac{V^T_B S_B}{s_{qB}}\Bigr) \nonumber \\ 
      &= \widetilde{X}
    \label{equ:quant-B}
\end{align}
\begin{equation}
    Y \approx s_{qW} s_{qX} Q\Bigl(\frac{W}{s_{qW}}\Bigr) Q\Bigl(\frac{\widetilde{X}}{s_{qX}}\Bigr) = \widetilde{Y}
    \label{equ:quant-Y}
\end{equation}
\begin{align}
    y &\approx s_{qA} s_{qA} s_{qY} Q\Bigl(\frac{V_A S_A}{s_{qA}}\Bigr) Q\Bigl(\frac{\widetilde{Y}}{s_{qY}}\Bigr) Q\Bigl(\frac{S_A V^T_A}{s_{qA}}\Bigr) \nonumber \\
      &= \widetilde{y}
    \label{equ:quant-y}
\end{align}
where $s_{q*}$ are the group-wise quantization scaling factors for the weights, activations, intermediate results, and Winograd transformation matrices and $Q$ is quantization function. 
We use simple min-max to dynamically quantize all activations during the forward pass. 

\subsection{Random Gaussian noise for fine-tuning scales}
We then use gradient descent (SGD) to optimize the following objective
and use the learned Winograd scale matrices $S_G$ and $S_B$ to determine the group-wise quantization scale factors, $s_{q*}$ in Eq.~\ref{equ:quant-B}-\ref{equ:quant-y}. 
For ease of setup, we treat each convolution layers independently.
Unlike previous studies that use domain-specific data for QAT or finetuning, we instead only use random Gaussian or random uniform noise to learn the Winograd scaling matrices.
In previous studies, the quantized model was calibrated and finetuned using a few samples from the training data, which might affect the quantized foundation diffusion models' generalization to unknown cases and downstream tasks. In addition, rather than finetuning scale factors separately for each convolution layer of a diffusion model, we learn a single set of finetuned scale factors for all layers. This further enhances the generalizability of our method.

\begin{equation}
    S^*_G, S^*_B = \arg \min_{S_G, S_B} \sum_{i \in D} \| y_i - \widetilde{y}_i \|
\end{equation}

%% file: sec/evaluation.tex
\section{Experiments}
\label{sec:evaluation}
\subsection{Experimental settings}
In this section, we evaluate the proposed learnable Winograd scales method on two latent diffusion models, InstaFlow-0.9B \cite{liu2024instaflowstephighqualitydiffusionbased} and Stable Diffusion \cite{rombach2022highresolutionimagesynthesislatent}, for text-to-image generation using the MS-COCO 2017 dataset \cite{lin2015microsoftcococommonobjects}, which contains $5000$ images. We use pre-trained checkpoints with a image resolution of $512 \times 512$, and for stable diffusion, we follow the setup in \cite{liu2024instaflowstephighqualitydiffusionbased} and choose the DPMSolver++ \cite{lu2023dpmsolverfastsolverguided} sampler with $25$ time steps and the classifier-free guidance scale of $5.0$. To further demonstrate the applicability of the proposed method, we also experiment with ResNets \cite{he2015deepresiduallearningimage} on image classification task with the Imagenet \cite{5206848} dataset. 

We apply group-wise quantization to all linear and convolution layers of all components of the diffusion model pipeline, including the text encoder, UNet, and decoder. We also quantize the attention weights, attention query and key multiplication, and value multiplication, which are not often performed in previous studies~\cite{Li_2023_ICCV}.
Furthermore, we compare the effects of two autoencoder models, the original autoencoder in \cite{rombach2022highresolutionimagesynthesislatent}, denoted as AKL, and the tiny autoencoder \cite{taesd}, denoted as TAESD.
To evaluate the quality of images generated, we followed the practices of previous works and computed FID \cite{heusel2018ganstrainedtimescaleupdate} and CLIP \cite{radford2021learningtransferablevisualmodels} (ViT-g-14 model) scores using the torchmetrics package. 

In the following, we first show results of applying only group-wise quantization and then the results of combining it with Winograd convolution using our proposed method. 

\begin{algorithm} [t]
\caption{Data-free fully group-wise quantized Winograd transform by scales optimization}\label{alg:finetune_scale}
\begin{algorithmic}[1]
    \Require Number of epochs $N$, number of batches per epoch $B$, number of convolution layers selected in each training iteration $K$, a collection of standard convolution layers with pre-trained weights $C_1$, a collection of corresponding group-wise quantized Winograd convolution layers $C_2$, and they all share a single set of learnable transformation scaling factors $S_G, S_B$. 
    \Ensure Learned scaling factors $S_G, S_B$ for all layers in $C_2$. 
    \For {$i = 1$ to $N$}
        \For { $j = 1$ to $B$}
            \State Randomly pick $K$ conv layers from $C_1$, e.g., 2 
            \State Find the corresponding $K$ Winograd conv layers from $C_2$ 
            \State Generate random noise inputs, Gaussian or uniform, $x_i$ for each pair of $Conv2d_i, WinoConv2d_i$
            \For { $k = 1$ to $K$ }
                \State $y_i = Conv2d_i(x_i)$
                \State $\tilde{y}_i = WinoConv2d_i(x_i) $
                \State loss = SQNR($y_i$, $\tilde{y}_i$)
                \State loss.backward()
            \EndFor 
            \State optimizer.step()
        \EndFor
    \EndFor
\end{algorithmic}
\end{algorithm}

\subsection{Text-to-image generation} 
Table \ref{tab:instaflow0.9b-akl} and table \ref{tab:instaflow0.9b-taesd} show the results for the InstaFlow-0.9B model with AKL and TAESD, respectively. It can be seen that with a bit width of W8A8, group-wise quantization maintains the same level of image generation quality and text-image alignment as the original FP16 model. For TAESD, it even improved upon the baseline FP16 model in both metrics. Quantizing the weights to $4$-bit leads to more distortion, but there is no significant degradation of image quality. When applying group-wise quantization to the standard Winograd convolution, 
FID and CLIP scores degrade drastically. Our method can restore most of the image generation quality for both $F(4, 3)$ and $F(6, 3)$ configurations.

\begin{figure}[t]
    \centering
    \includegraphics[width=0.96\linewidth]
    {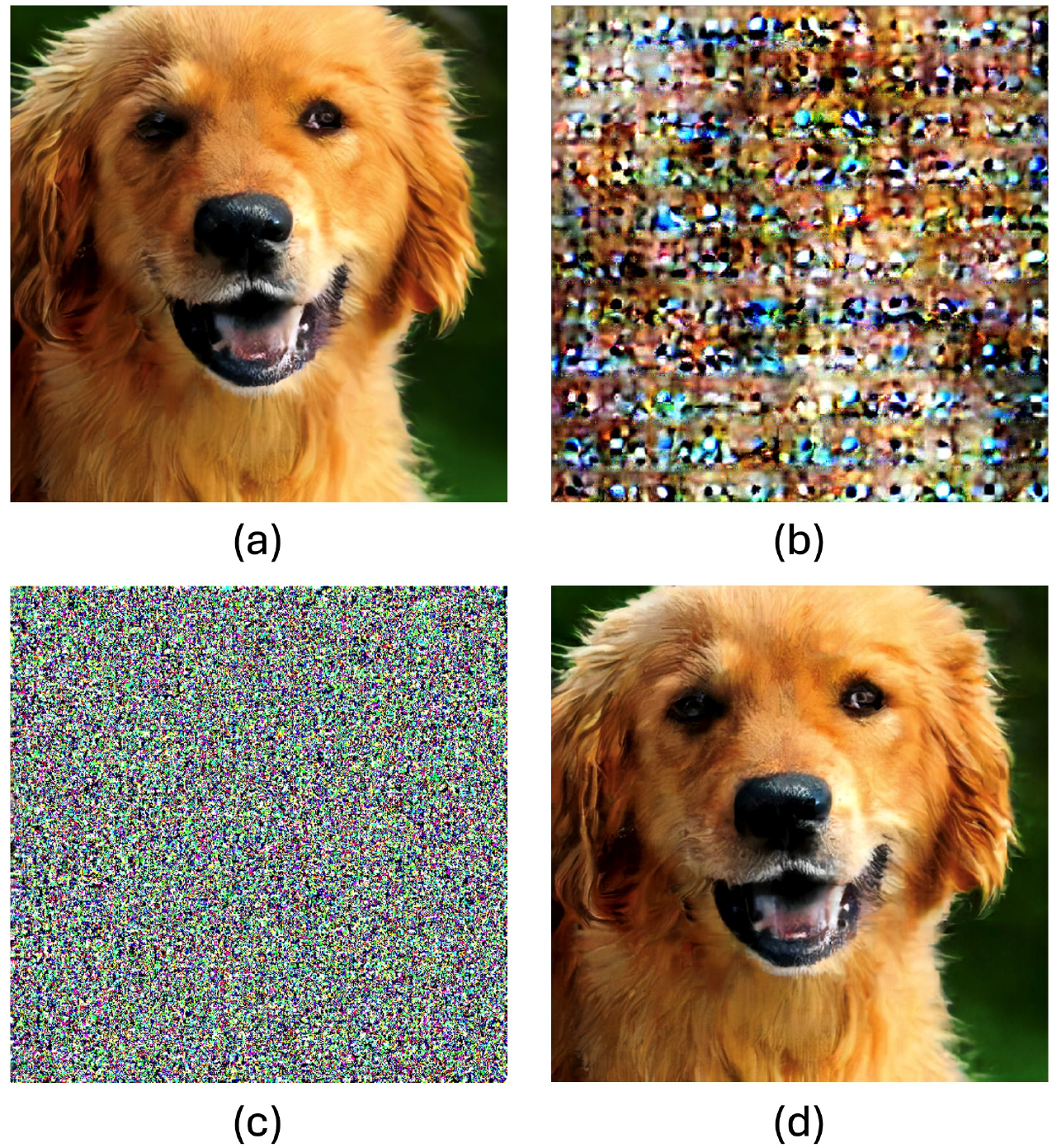}
    \caption{(a) Image generated from FP16 model with AKL. (b) (c) Images generated from W8A8 group-wise quantized standard Winograd convolution, using AKL and TAESD, respectively. (d) Image generated from W8A8 group-wise quantized Winograd convolution with learned scales and AKL. InstaFlow-0.9B model and Winograd F(6, 3) was used.}
    \label{fig:dog-image}
\vspace{1mm}
\end{figure}

\begin{figure}[t]
    \centering
    \includegraphics[width=0.96\linewidth]
    {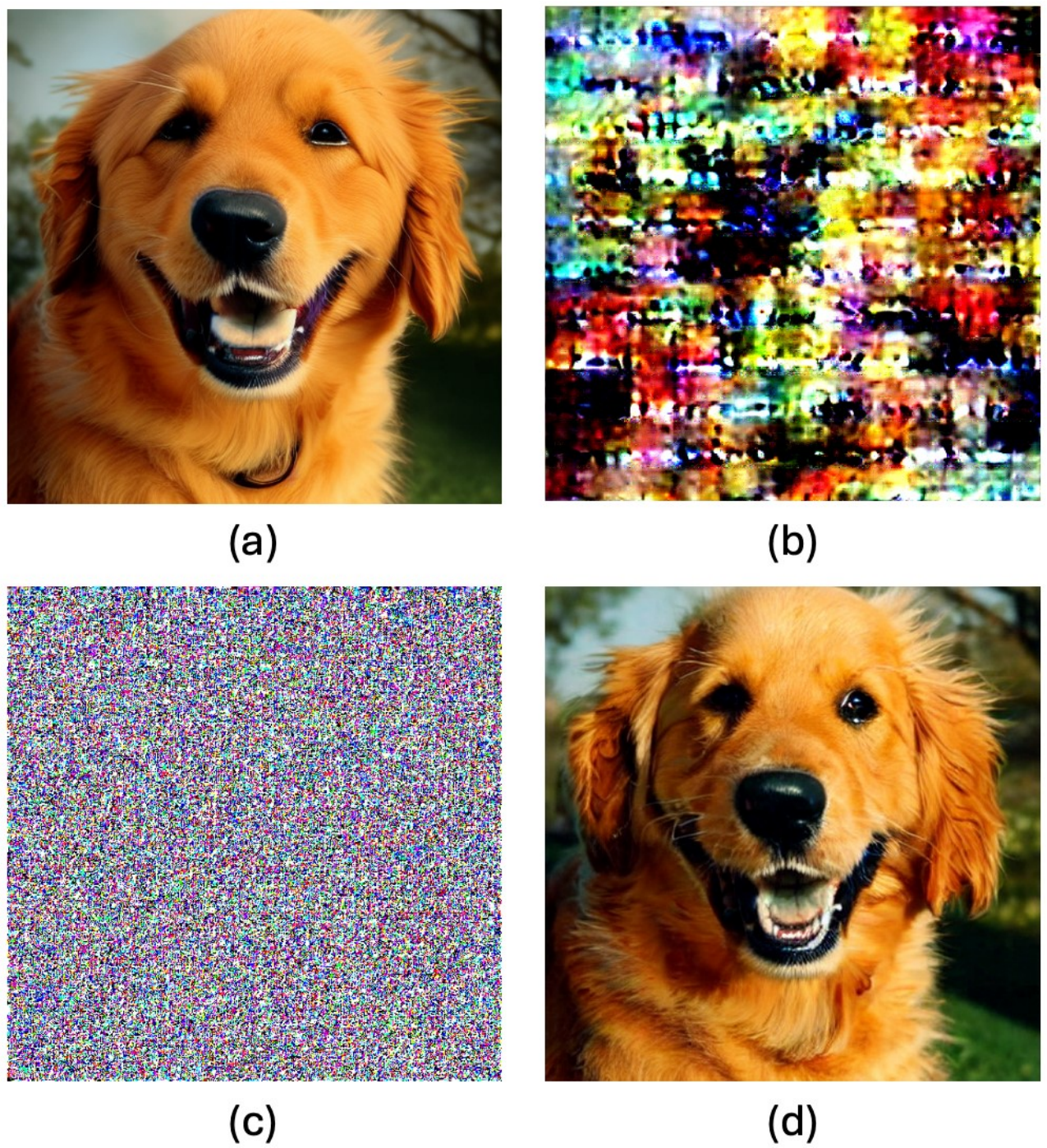}
    \caption{(a) Image generated from FP16 model with AKL. (b) (c) Images generated from W8A8 group-wise quantized standard Winograd convolution, using AKL and TAESD, respectively. (d) Image generated from W8A8 group-wise quantized Winograd convolution with learned scales and AKL. Stable Diffusion V1.5 model and Winograd F(6, 3) was used.}
    \label{fig:dog-image-sd}
\end{figure}

Figure~\ref{fig:dog-image} and Figure \ref{fig:dog-image-sd} show some qualitative examples of the challenges of direct application of group-wise quantization to the Winograd pipeline, especially for TAESD where the images are completely destroyed. Because of this, we do not compute FID and CLIP scores for TAESD under this condition. As shown in Figure \ref{fig:winograd_dist}, this is mainly due to the huge dynamic range differences in the Winograd output $Y$. After applying the learned Winograd scales obtained from Algorithm \ref{alg:finetune_scale}, both images are mostly recovered. Figure~\ref{fig:winograd_dist}(c) also shows the effectiveness of our method in dynamic range equalization in the Winograd output $Y$.

\vspace{-2mm}

\begin{table} [h!]
    \caption{Results on the InstaFlow-0.9B model with group-wise quantization, Winograd convolution, and AKL autoencoder.}
    \begin{small}
    \centering
    \begin{tabular}{lcccc}
    \toprule
        \multicolumn{5}{c}{IF-0.9B, COCO2017-5k, AKL} \\
        \hline
        Model & tile size & Bits & FID($\downarrow$) & CLIP($\uparrow$) \\
        \hline
        FP16 & N/A & 16/16 & 23.00 & 30.19 \\
        \hline
        W4A8 & N/A & 4/8 & 28.73 & 29.09 \\
        \hline
        W8A8 & N/A & 8/8 & 23.04 & 30.16 \\
        \hline
        W8A8 Winograd & F(4,3) & 8/8 & 217.16 & 15.14 \\\cline{2-5}
        Standard scales & F(6,3) & 8/8 & 326.96 & 5.95 \\
        \hline
        W8A8 Winograd & F(4,3) & 8/8 & 24.51 & 29.87 \\\cline{2-5}
        Learned scales & F(6,3) & 8/8 & 26.58 & 29.65 \\
        \bottomrule
    \end{tabular}
    \end{small}
    \label{tab:instaflow0.9b-akl}
\end{table}

\vspace{1.7mm}

\begin{table} [h!]
    \caption{Results on the InstaFlow-0.9B model with group-wise quantization, Winograd convolution, and TEASD autoencoder.}
    \centering
    \begin{small}
    \begin{tabular}{lcccc}
    \toprule
        \multicolumn{5}{c}{IF-0.9B, COCO2017-5k, TEASD} \\
        \hline
        Model & tile size & Bits & FID($\downarrow$) & CLIP($\uparrow$) \\
        \hline
        FP16 & N/A & 16/16 & 26.39 & 29.73 \\
        \hline
        W4A8 & N/A & 4/8 & 31.94 & 28.80 \\
        \hline
        W8A8 & N/A & 8/8 & 24.67 & 30.08 \\
        \hline
        W8A8 Winograd & F(4,3) & 8/8 & N/A & N/A \\\cline{2-5}
        Standard scales & F(6,3) & 8/8 & N/A & N/A \\
        \hline
        W8A8 Winograd & F(4,3) & 8/8 & 27.09 & 29.53  \\\cline{2-5}
        Learned scales & F(6,3) & 8/8 & 27.69 & 29.46 \\
        \bottomrule
    \end{tabular}
    \end{small}
    \label{tab:instaflow0.9b-taesd}
\end{table}

\vspace{1.2mm}

\begin{table} [h!]
    \caption{Results on the Stable Diffusion V1.5 model with group-wise quantization, Winograd convolution, and AKL autoencoder. }
    \centering
    \begin{small}
    \begin{tabular}{lcccc}
    \toprule
         \multicolumn{5}{c}{SD-1.5, COCO2017-5k, AKL} \\
        \hline
        Model & tile size & Bits & FID($\downarrow$) & CLIP($\uparrow$) \\
        \hline
        FP16 & N/A & 16/16 & 21.72 & 31.72 \\
        \hline
        W4A8 & N/A & 4/8 & 20.62 & 31.16 \\
        \hline
        W8A8 & N/A & 8/8 & 21.80 & 31.70 \\
        \hline
        W8A8 Winograd & F(4,3) & 8/8 & 312.03 & 7.22 \\\cline{2-5}
        Standard scales & F(6,3) & 8/8 & 329.25 & 3.56 \\
        \hline
        W8A8 Winograd & F(4,3) & 8/8 & 19.65 & 31.63 \\\cline{2-5}
        Learned scales & F(6,3) & 8/8 & 20.52 & 31.53 \\
        \bottomrule
    \end{tabular}
    \end{small}
    \label{tab:sdv15-akl}
\end{table}
\vspace{3mm}

\begin{table} [h!]
    \caption{Results on the Stable Diffusion V1.5 model with group-wise quantization, Winograd convolution, and TEASD autoencoder.} 
    \centering
    \begin{small}
    \begin{tabular}{lcccc}
    \toprule
         \multicolumn{5}{c}{SD-1.5, COCO2017-5k, TEASD} \\
        \hline
        Model & tile size & Bits & FID($\downarrow$) & CLIP($\uparrow$) \\
        \hline
        FP16 & N/A & 16/16 & 22.04 & 31.76 \\
        \hline
        W4A8 & N/A & 4/8 & 21.78 & 31.12 \\
        \hline
        W8A8 & N/A & 8/8 & 22.10 & 31.74 \\
        \hline
        W8A8 Winograd & F(4,3) & 8/8 & N/A & N/A \\\cline{2-5}
        Standard scales & F(6,3) & 8/8 & N/A & N/A \\
        \hline
        W8A8 Winograd & F(4,3) & 8/8 & 21.47 & 31.31 \\\cline{2-5}
        Learned scales & F(6,3) & 8/8 & 20.56 & 31.55 \\
        \bottomrule
    \end{tabular}
    \end{small}
    \label{tab:sdv15-taesd}
\vskip -0.1in
\end{table}

\begin{table*}[!thb]
    \caption{ResNets Top-1 accuracy results on the ImageNet dataset.}
    \centering
    \begin{small}
    \begin{tabular}{cccccccc}
         \toprule
         Model & tile size & Bits & Standard Winograd & BQW \cite{bqw} & PAW+FSQ \cite{chen2024towards} & Winograd with Learned Scales \\
         \hline
         ResNet18 &  F(4,3) & W8/A8 & 4.26\% & 68.94\% & 68.16\% & \textbf{69.44\%} \\
         69.76\% &   F(6,3) & W8/A8 & 0.08\% & 66.08\% & 66.89\% & \textbf{68.51\%} \\
         \hline
         ResNet34 &  F(4,3) & W8/A8 & 5.98\% & 72.86\% & 71.75\% & \textbf{73.06}\% \\
         73.30\% &  F(6,3) & W8/A8 & 0.18\% & 70.87\% & 69.72\% & \textbf{72.28\%}\\
         \hline
         ResNet50 &  F(4,3) & W8/A8 & 56.84\% & \textbf{76.10}\% & 73.84\% & 75.87\% \\
         76.15\% &  F(6,3) & W8/A8 & 29.45\% & \textbf{75.79}\% & 75.36\% & 75.11\% \\
         \bottomrule
    \end{tabular}
    \end{small}
    \label{tab:resnets-results}
\end{table*}

Table \ref{tab:sdv15-akl} and Table \ref{tab:sdv15-taesd} show the evaluation results from Stable diffusion V1.5, using DPMSolver++ \cite{lu2023dpmsolverfastsolverguided} sampler with $25$ sampling steps and a classifier-free guidance scale of $5.0$. Similar to InstaFlow-0.9B, we observe that group-wise quantization to W8A8 exhibited almost no degradation in image generation quality with either AKL or TEASD. Direct application of group-wise quantization to Winograd convolution also leads to almost complete loss of model quality, e.g., FID increases from $21.72$ to $329.25$, but this is reversed to the most extent by our method. It is worth noting that, while the bit-width setting of W4A8 and some Winograd cases showed improvements in FID when compared to the baseline FP16 model, we observe some distortions in some of the generated images when compared to the W8A8 quantization or the FP16 model. This may have to do with the unreliability of the FID metric in image quality evaluation, as mentioned in \cite{naeem2020reliablefidelitydiversitymetrics, jayasumana2024rethinkingfidbetterevaluation, stein2023exposingflawsgenerativemodel}. 

\subsection{Image classification}
The evaluation results from ResNets on the ImageNet 
are shown in Table \ref{tab:resnets-results}. We 
compare it to two methods: BQW~\cite{bqw} and PAW+FSQ~\cite{chen2024towards}. Similar to findings from the diffusion models, group-wise quantization of the standard Winograd convolution 
to W8A8 significantly degraded ResNets classification accuracy in both $F(4, 3)$ and $F(6, 3)$ configurations. After utilizing the learned Winograd transform scales, we recovered most of the accuracies of the original full-precision model. 
Furthermore, our method outperformed PAW+FSQ by $1.62\%$ and $2.56\%$ in top-1 ImageNet accuracy on ResNet-18 and ResNet-34, respectively, for Winograd F(6, 3).

\subsection{Runtime improvements}
\begin{figure}
    \centering
    \includegraphics[width=0.75\linewidth]
    {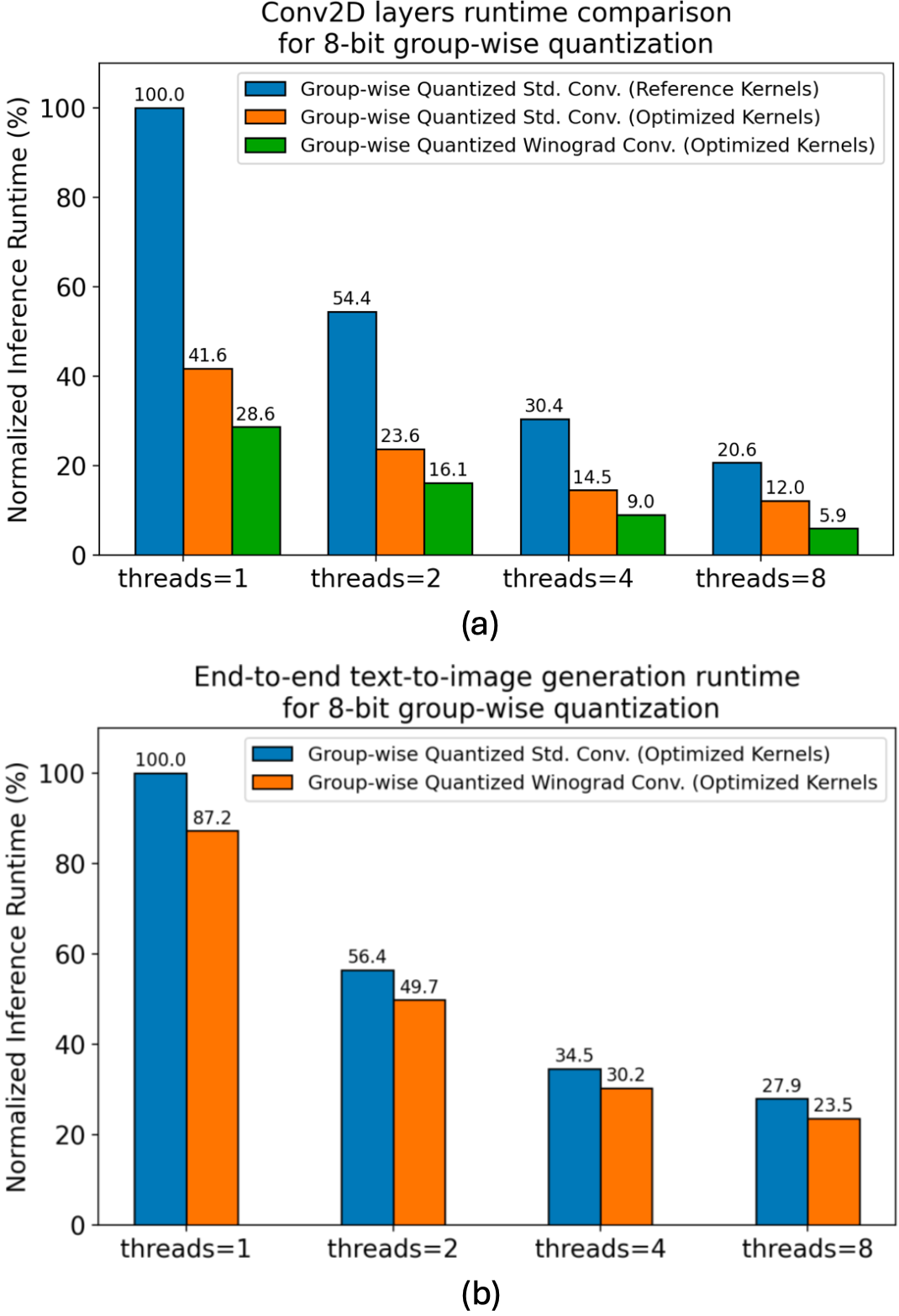}
    \caption{Convolution layers runtime improvements by using group-wise fully-quantized Winograd convolution.}
    \label{fig:conv-runtime}
\end{figure}

Finally, we measure the runtime improvement of convolution layers as well as the end-to-end runtime of the InstaFlow-0.9B diffusion model on Arm Graviton3 CPUs with varying thread counts (cores). 
We use the stable-diffusion.cpp \cite{sdcpp} framework to collect the inference runtime.
As evident from Figure~\ref{fig:conv-runtime}(a), the highly optimized kernels offer a significant improvement in runtime for group-wise quantized convolution layers. The fully-quantized Winograd convolution provides an additional $31.3\%$ relative improvement over the optimized standard convolution layers. This translates into an $12.8\%$ overall runtime improvement for the diffusion model for a single thread, as shown in Figure~\ref{fig:conv-runtime}(b).

%% file: sec/conclusion.tex
\section{Conclusion}
\label{sec:conclusion}
The exorbitant computational and storage overhead of large-scale diffusion models limit their practicality for on-device inference. We present a novel, lightweight method that (1) employs group-wise PTQ on both weights and activations, and (2) fully quantizes fast Winograd convolution by fine-tuning only the scale parameters of the transformation matrices, all without requiring any calibration data. Our method, implemented using highly optimized CPU kernels, maintains image generation quality and offers nearly full-precision image classification accuracy while significantly speeding up CPU wall-clock time for large-scale diffusion models.

%% file: supplementary_arxiv.tex








\clearpage
\maketitlesupplementary

\section{Winograd transformations for convolution}
\subsection{Standard Winograd transforms}
\label{sec:winograd_matrices}
Following \cite{vincent2017improving}, given a set of polynomial points $(f_i, g_i)$, the Vandermonde matrix $V_{a \times b}$ is constructed as below, 

\begin{equation}
\begin{bmatrix}
    f_0^0 g_0^{b-1} & f_0^1 g_0^{b-2} & \cdots & f_0^{b-1} g_0^0 \\
    f_1^0 g_1^{b-1} & f_1^1 g_1^{b-2} & \cdots & f_1^{b-1} g_1^0 \\
    \vdots & \vdots & \ddots & \vdots \\
    f_{a-1}^0 g_{a-1}^{b-1} & f_{a-1}^1 g_{a-1}^{b-2} & \cdots & f_{a-1}^{b-1} g_{a-1}^0 \\
\end{bmatrix}
\end{equation}

For standard Winograd convolution, we adopt the widely used polynomial points and scaling factors, as mentioned in \cite{lavin2015fastalgorithmsconvolutionalneural, bqw, vincent2017improving, alam2022winogradconvolutiondeepneural}. Specifically, for $F(4, 3)$ the polynomial points, scaling factors, Vandermonde and transformation matrices are the following, 
\begin{align}
    (f_i, g_i) = &[ (0, 1), (1, 1), (-1, 1), \nonumber \\ 
    &(2, 1), (-2, 1), (1, 0) ] \nonumber  \\
    S_A = &[1, 1, 1, 1, 1, 1]  \\
    S_B = &[4, -6, -6, 24, 24, 1]  \nonumber \\
    S_G = &[\nicefrac{1}{4}, -\nicefrac{1}{6}, -\nicefrac{1}{6}, \nicefrac{1}{24}, \nicefrac{1}{24}, 1]  \nonumber
\end{align}

\begin{equation}
V_{6\times4} = 
\begin{bmatrix*}[r]
    1 & 0 & 0 & 0 \\
    1 & 1 & 1 & 1 \\ 
    1 & -1 & 1 & -1 \\ 
    1 & 2 & 4 & 8 \\
    1 & -2 & 4 & -8 \\
    0 & 0 & 0 & 1
\end{bmatrix*}
\end{equation}

\begin{equation}
V_{6\times6}^{-T} = 
\begin{bmatrix*}[r]
    1 & 0 & -\nicefrac{5}{4} & 0 & \nicefrac{1}{4} & 0 \\
    0 & \nicefrac{2}{3} & \nicefrac{2}{3} & -\nicefrac{1}{6} & -\nicefrac{1}{6} & 0 \\ 
    0 & -\nicefrac{2}{3} & \nicefrac{2}{3} & \nicefrac{1}{6} & -\nicefrac{1}{6} & 0 \\ 
    0 & -\nicefrac{1}{12} & -\nicefrac{1}{24} & \nicefrac{1}{12} & \nicefrac{1}{24} & 0 \\ 
    0 & \nicefrac{1}{12} & -\nicefrac{1}{24} & -\nicefrac{1}{12} & \nicefrac{1}{24} & 0 \\ 
    0 & 4 & 0 & -5 & 0 & 1 \\
\end{bmatrix*}
\end{equation}

\begin{equation}
V_{6\times3} = 
\begin{bmatrix*}[r]
    1 & 0 & 0 \\
    1 & 1 & 1 \\ 
    1 & -1 & 1 \\ 
    1 & 2 & 4 \\
    1 & -2 & 4 \\
    0 & 0 & 1
\end{bmatrix*}
\end{equation}

\begin{align}
A^T &= V_{6\times4}^T diag(S_A) \nonumber \\
&= 
\begin{bmatrix*}[r]
    1 & 1 & 1 & 1 & 1 & 0 \\
    0 & 1 & -1 & 2 & -2 & 0  \\ 
    0 & 1 & 1 & 4 & 4 & 0  \\ 
    0 & 1 & -1 & 8 & -8 & 1 
\end{bmatrix*}
\end{align}

\begin{align}
B^T &= diag(S_B) V_{6\times6}^{-T} \nonumber \\
&=
\begin{bmatrix*}[r]
     4 & 0 & -5 & 0 & 1 & 0 \\
     0 & -4 & -4 & 1 & 1 & 0 \\
     0 & 4 & -4 & -1 & 1 & 0 \\
     0 & -2 & -1 & 2 & 1 & 0 \\
     0 & 2 & -1 & -2 & 1 & 0 \\
     0 & 4 & 0 & -5 & 0 & 1 
\end{bmatrix*}
\end{align}

\begin{align}
G &= diag(S_G) V_{6\times3} \nonumber \\
&= 
\begin{bmatrix*}[r]
    \nicefrac{1}{4}  & 0 & 0  \\
    -\nicefrac{1}{6} & -\nicefrac{1}{6} & -\nicefrac{1}{6} \\ 
    -\nicefrac{1}{6} & \nicefrac{1}{6} & -\nicefrac{1}{6} \\ 
    \nicefrac{1}{24} & \nicefrac{1}{12} & \nicefrac{1}{6}  \\ 
    \nicefrac{1}{24} & -\nicefrac{1}{12} & \nicefrac{1}{6}  \\ 
    0 & 0 & 1 \\ 
\end{bmatrix*}
\end{align}

For $F(6, 3)$, the polynomial points, scaling factors, Vandermonde and transformation matrices are the following, 
\begin{align}
(f_i, g_i) = &[ (0, 1), (1, 1), (-1, 1), (2, 1),  \nonumber \\ 
        &(-2, 1), (\nicefrac{1}{2}, 1), (-\nicefrac{1}{2}, 1), (1, 0) ] \nonumber  \\
    S_A = &[1, 1, 1, 1, 1, 1, 1, 1]  \\
    S_B = &[1, -\nicefrac{9}{2}, -\nicefrac{9}{2}, 90, 90, \nicefrac{45}{32}, \nicefrac{45}{32}, 1] 
 \nonumber \\
    S_G = &[1, -\nicefrac{2}{9}, -\nicefrac{2}{9}, \nicefrac{1}{90}, \nicefrac{1}{90}, \nicefrac{32}{45}, \nicefrac{32}{45}, 1]  \nonumber
\end{align}

\begin{equation}
V_{8\times6} = 
\begin{bmatrix*}[r]
    1 & 0 & 0 & 0 & 0 & 0 \\
    1 & 1 & 1 & 1 & 1 & 0 \\
    1 & -1 & 1 & -1 & 1 & -1 \\
    1 & 2 & 4 & 8 & 16 & 32 \\
    1 & -2 & 4 & -8 & 16 & -32 \\
    1 & \nicefrac{1}{2} & \nicefrac{1}{4} & \nicefrac{1}{8} & \nicefrac{1}{16} & \nicefrac{1}{32} \\
    1 & -\nicefrac{1}{2} & \nicefrac{1}{4} & -\nicefrac{1}{8} & \nicefrac{1}{16} & -\nicefrac{1}{32} \\
    0 & 0 & 0 & 0 & 0 & 1 \\
\end{bmatrix*}
\end{equation}

\begin{equation}
V_{8\times8}^{-T} = 
\begin{bmatrix*}
    1 & 0 & \nicefrac{-21}{4} & 0 & \nicefrac{21}{4} & 0 & -1 & 0 \\
    0 & -\nicefrac{2}{9} & -\nicefrac{2}{9} & \nicefrac{17}{18} & \nicefrac{17}{18} & -\nicefrac{2}{9} & -\nicefrac{2}{9} & 0 \\ 
    0 & \nicefrac{2}{9} & -\nicefrac{2}{9} & -\nicefrac{17}{18} & -\nicefrac{17}{18} & \nicefrac{2}{9} & -\nicefrac{2}{9} & 0 \\ 
    0 & \nicefrac{1}{180} & \nicefrac{1}{360} & -\nicefrac{1}{36} & -\nicefrac{1}{72} & \nicefrac{1}{45} & \nicefrac{1}{90} & 0 \\ 
    0 & -\nicefrac{1}{180} & \nicefrac{1}{360} & \nicefrac{1}{36} & -\nicefrac{1}{72} & -\nicefrac{1}{45} & \nicefrac{1}{90} & 0 \\ 
    0 & \nicefrac{64}{45} & \nicefrac{128}{45} & -\nicefrac{16}{9} & -\nicefrac{32}{9} & \nicefrac{16}{45} & \nicefrac{32}{45} & 0\\ 
    0 & -\nicefrac{64}{45} & \nicefrac{128}{45} & \nicefrac{16}{9} & -\nicefrac{32}{9} & -\nicefrac{16}{45} & \nicefrac{32}{45} & 0\\ 
    0 & -\nicefrac{1}{4} & 0 & \nicefrac{21}{4} & 0 & -\nicefrac{21}{4} & 0 & 1\\ 
\end{bmatrix*}
\end{equation}

\begin{equation}
V_{8\times3} = 
\begin{bmatrix*}
    1 & 0 & 0  \\
    1 & 1 & 1  \\
    1 & -1 & 1 \\
    1 & 2 & 4  \\
    1 & -2 & 4 \\
    1 & \nicefrac{1}{2} & \nicefrac{1}{4} \\
    1 & -\nicefrac{1}{2} & \nicefrac{1}{4} \\
    0 & 0 & 1 \\
\end{bmatrix*}
\end{equation}

\begin{align}
A^T &= V_{8\times6}^T diag(S_A) \nonumber \\
&= 
\begin{bmatrix*}[r]
    1 & 1 & 1 & 1 & 1 & 1 & 1 & 0 \\
    0 & 1 & -1 & 2 & -2 & \nicefrac{1}{2} & -\nicefrac{1}{2} & 0  \\ 
    0 & 1 & 1 & 4 & 4 & \nicefrac{1}{4} & \nicefrac{1}{4} & 0  \\ 
    0 & 1 & -1 & 8 & -8 & \nicefrac{1}{8} & -\nicefrac{1}{8} & 0 \\
    0 & 1 & 1 & 16 & 16 & \nicefrac{1}{16} & \nicefrac{1}{16} & 0 \\
    0 & 1 & -1 & 32 & -32 & \nicefrac{1}{32} & -\nicefrac{1}{32} & 0  \\
\end{bmatrix*}
\end{align}

\begin{align}
B^T &= diag(S_B) V_{8\times8}^{-T} \nonumber \\
&=
\begin{bmatrix*}[r]
    4 & 0 & -21 & 0 & 21 & 0 & -4 & 0 \\
    0 & 4 & 4 & -17 & -17 & 4 & 4 & 0 \\ 
    0 & -4 & 4 & 17 & -17 & -4 & 4 & 0 \\ 
    0 & 2 & 1 & -10 & -5 & 8 & 4 & 0 \\ 
    0 & -2 & 1 & 10 & -5 & -8 & 4 & 0 \\ 
    0 & 8 & 16 & -10 & -20 & 2  & 4 & 0\\ 
    0 & -8 & 16 & 10 & -20 & -2 & 4 & 0\\ 
    0 & -4 & 0 & 21 & 0 & -21 & 0 & 4
\end{bmatrix*}
/4  
\end{align}

\begin{align}
G &= diag(S_G) V_{6\times3} \nonumber \\
&=
    \begin{bmatrix*}[r]
        1 & 0 & 0 \\
        -\nicefrac{2}{9} & -\nicefrac{2}{9} & -\nicefrac{2}{9} \\
        -\nicefrac{2}{9} & \nicefrac{2}{9} & -\nicefrac{2}{9} \\
        \nicefrac{1}{90} & \nicefrac{1}{45} & \nicefrac{2}{45} \\
        \nicefrac{1}{90} & -\nicefrac{1}{45} & \nicefrac{2}{45} \\
        \nicefrac{32}{45} & \nicefrac{16}{45} & \nicefrac{8}{45} \\
        \nicefrac{32}{45} & -\nicefrac{16}{45} & \nicefrac{8}{45} \\
        0 & 0 & 1
    \end{bmatrix*}
\end{align}

\subsection{Learned Winograd scales}
Table~\ref{tab:learned-scales} shows the difference in values between the standard Winograd scales and our learned Winograd scales for $F(6, 3)$. It is worth noting that the magnitudes of $S_A$ have become smaller while those of $S_G$ are bigger, and $S_B$ stays relatively unchanged. 

\begin{table}[t]
    \caption{Comparison between standard Winograd scales and our learned Winograd scales.}
    \begin{small}
    \centering
    \begin{tabular}{lcccccc}
        \toprule
         tile size & \multicolumn{3}{c}{Standard Winograd scales} & \multicolumn{3}{c}{Learned Winograd} scales  \\ 
         \hline 
         \multirow{9}{*}{$F(6,3)$} & $S_A$ & $S_B$ & $S_G$ & $S_A$ & $S_B$ & $S_G$ \\
         \cline{2-7} 
                                   & 1 & 1 & 1 & 0.525 & -1.378 & -1.382 \\
                                   & 1 & -4.5 & -0.222 & 0.354 & -4.908 & -0.576 \\ 
                                   & 1 & -4.5 & -0.222 & 0.351 & -4.912 & -0.579 \\
                                   & 1 & 90 & 0.0111 & 0.0601 & 90.366 & 0.184 \\
                                   & 1 & 90 & 0.0111 & 0.0609 & 90.362 & 0.182 \\
                                   & 1 & 1.406 & 0.711 & 0.491 & 1.820 & 1.119 \\
                                   & 1 & 1.406 & 0.711 & 0.490 & 1.823 & 1.120 \\
                                   & 1 & 1 & 1 & 0.519 & 1.386 & 1.391 \\
         \bottomrule
    \end{tabular}
    \end{small}
    \label{tab:learned-scales}
\end{table}

\begin{table} [h!]
    \caption{Results on the InstaFlow-0.9B model with group-wise quantization, Winograd convolution, and AKL autoencoder. Comparison between learning scales and learning transformation matrices. }
    \begin{small}
    \centering
    \begin{tabular}{lcccc}
        \toprule
        \multicolumn{5}{c}{IF-0.9B, COCO2017-5k, AKL} \\
        \hline
        Model & tile size & Bits & FID($\downarrow$) & CLIP($\uparrow$) \\
        \hline
        FP16 & N/A & 16/16 & 23.00 & 30.19 \\
        \hline
        W4A8 & N/A & 4/8 & 28.73 & 29.09 \\
        \hline
        W8A8 & N/A & 8/8 & 23.04 & 30.16 \\
        \hline
        W8A8 Winograd & F(4,3) & 8/8 & 217.16 & 15.14 \\\cline{2-5}
        Standard scales & F(6,3) & 8/8 & 326.96 & 5.95 \\
        \hline
        W8A8 Winograd & F(4,3) & 8/8 & 24.51 & 29.87 \\\cline{2-5}
        Learned scales & F(6,3) & 8/8 & 26.58 & 29.65 \\
        \hline
        W8A8 Winograd & F(4,3) & 8/8 & 204.00 & 17.13 \\\cline{2-5}
        Learned transforms & F(6,3) & 8/8 & 371.86 & 3.38 \\
        \bottomrule
    \end{tabular}
    \end{small}
    \label{tab:instaflow0.9b-akl-learn-transform}
\end{table}

\section{Comparison with learning transformation matrices instead of Winograd scales} 
\cite{fernandez2020searching} proposed to treat the Winograd transformation matrices $A$, $B$, and $G$ as learnable parameters and jointly optimize them with other model weights and biases in a QAT setup. Although this is much less practical in the domain of Generative AI, as mentioned above, we still adopt this paradigm to compare with our method. All training setups are the same except that transformation matrices $A$, $B$, and $G$ are learned directly using random noise inputs instead of 
learning only the scaling factors $S_A$, $S_B$, and $S_G$. The results are shown in Table \ref{tab:instaflow0.9b-akl-learn-transform} using the InstaFlow-0.9B model with AKL autoencoder. It can be seen that learning the transformation matrices directly offers almost no improvement compared to the standard Winograd transforms. This could be due to the use of random noise as layer inputs and treating each layer independently.

\section{Advantage of data-free approach over calibration data and other Winograd methods}
Using our paradigm, we fine-tuned Winograd scales and transformation matrices in both end-to-end and BRECQ \cite{li2021brecqpushinglimitposttraining} modes. We compute loss using the difference between the generated images or features of the fully group-wise quantized model and its FP16 counterpart. 
We randomly select 10k prompts from the poloclub/diffusiondb \cite{wangDiffusionDBLargescalePrompt2022} dataset for calibration and use the same setup as when training with random noise. It is more challenging to set up an end-to-end training pipeline, and the diffusion model as a whole needs to be kept on the GPU, hence requiring longer training time and more memory. It can be seen in Table \ref{tab:train-e2e} that learning scales with calibration data is also effective for $F(4, 3)$, but less so for $F(6,3)$ than learning with noise. 
The poor accuracy for calibration data when compared to our method may be attributed to the small sample size of 10k prompts, but using more samples may result in overfitting. Similar to data-free training, learning Winograd transforms with calibration data failed to produce good-quality generations. Furthermore, training with BRECQ takes significantly longer ($>10\times$) due to a much more complex training pipeline, with each layer having its own trained Winograd scales, as opposed to a single set of Winograd scales for all layers of one network in our method. 

Channel balancing (BQW) \cite{bqw} only quantizes the Hadamard product, which we could achieve using only group-wise quantization. PAW+FSQ \cite{chen2024towards} is perhaps not suitable for large-scale diffusion models: transformation matrices are difficult to finetune with the use of calibration data, as shown in Table ~\ref{tab:train-e2e}. The use of calibration data for PAW can make it difficult to ensure generalizability to unseen downstream tasks for foundation models. FSQ optimizes hadamard product output (Y) quantization, whereas our learned scales jointly optimize the entire Winograd algorithm, including transformation matrices and Y.

\section{Comparison of group-wise quantization against other quantization methods} 
Table \ref{tab:sdv14-akl-qdiffusion-dpm} and Table \ref{tab:sdv14-akl-qdiffusion-plms} show the comparison between our group-wise quantization method against a popular, recently proposed quantization scheme, called Q-Diffusion\cite{Li_2023_ICCV}, for Stable Diffusion V1.4\cite{rombach2022highresolutionimagesynthesislatent} with DPMSolver++\cite{lu2023dpmsolverfastsolverguided} and PLMS\cite{liu2022pseudonumericalmethodsdiffusion} sampler, respectively. All models were sampled for $25$ steps, and MSCOCO 2017 was used to generate FID and CLIP scores. We conducted the experiments with Q-Diffusion using the official codebase and pre-calibrated quantized checkpoints released by the authors. \\

\begin{figure*}[t]
    \centering
    \includegraphics[width=0.75\linewidth]{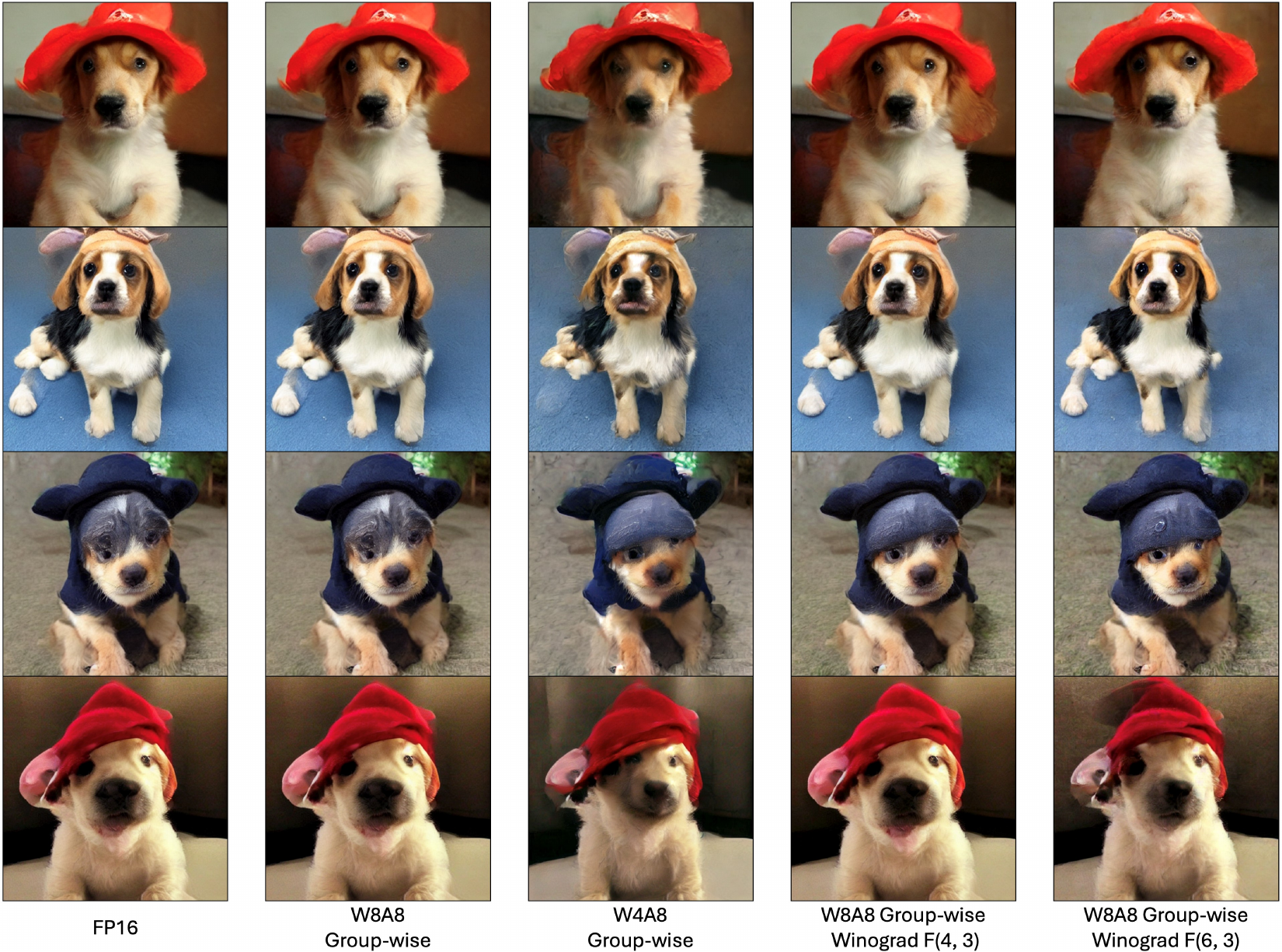}
    \caption{InstaFlow-0.9B with AKL. Prompt \textit{"A puppy wearing a hat; realistic"}}
    \label{fig:dogs-instaflow-comparison}
\end{figure*}

\begin{figure*}[h]
    \centering
    \includegraphics[width=0.75\linewidth]{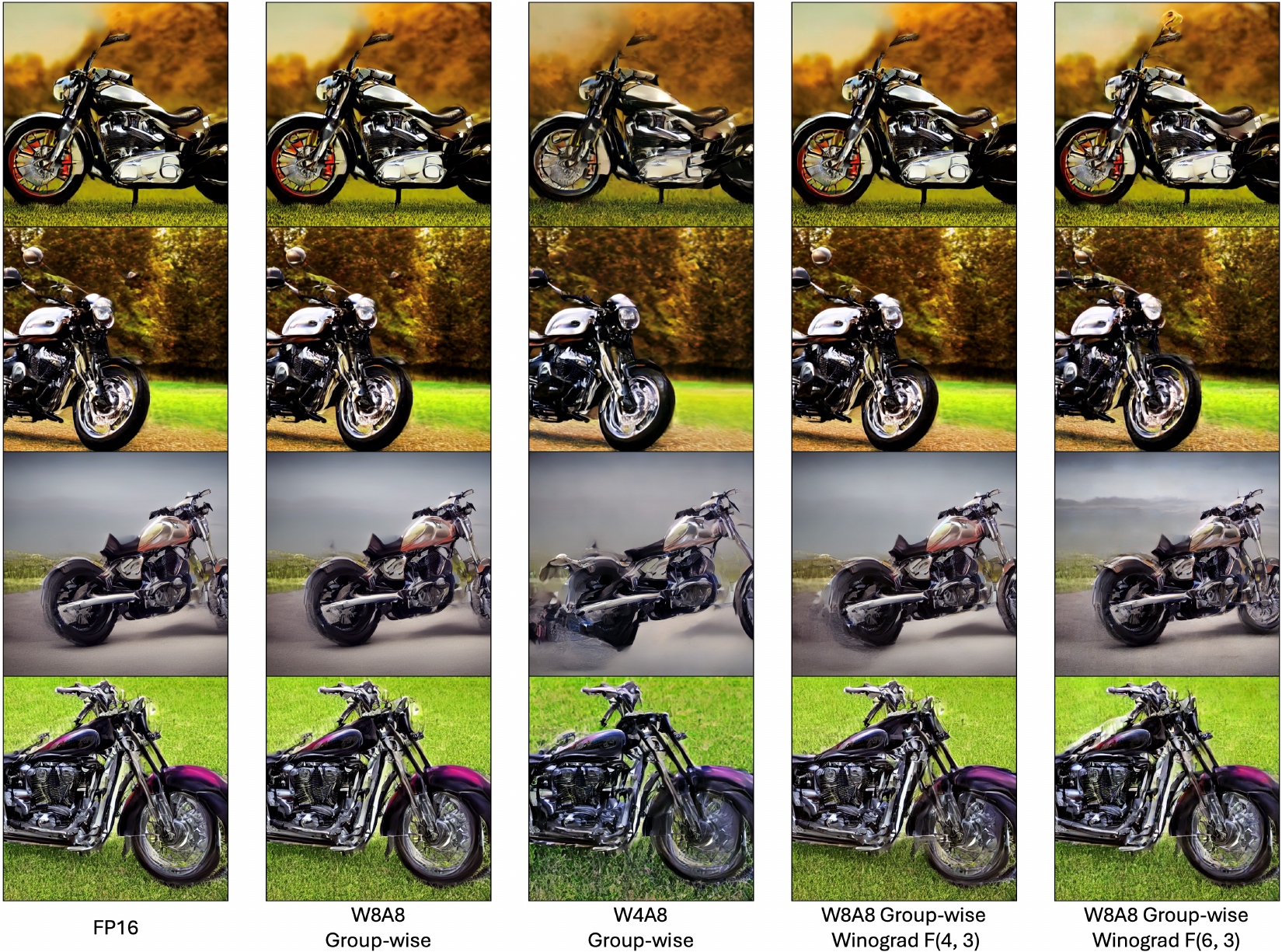}
    \caption{InstaFlow-0.9B with AKL. Prompt \textit{"A shiny motorcycle on the field; realistic, high-resolution"}}
    \label{fig:motor-instaflow-comparison}
\end{figure*}

\begin{figure*}[t]
    \centering
    \includegraphics[width=0.75\linewidth]{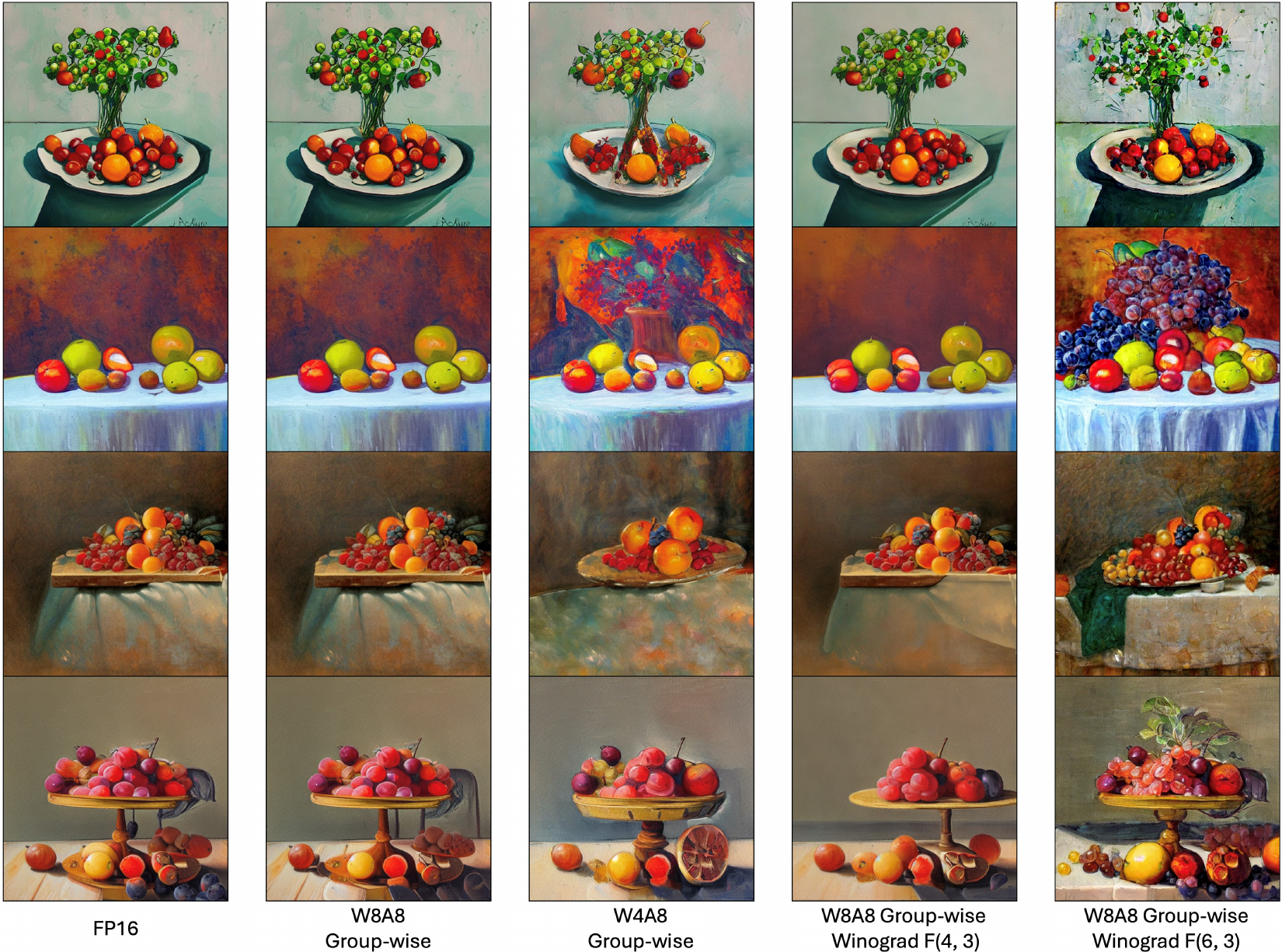}
    \caption{Stable Diffusion V1.5 with AKL and DPMSolver++ sampler. Prompt \textit{"A painting of a table with fruit on top of it"}}
    \label{fig:painting-sdv1.5-comparison}
\end{figure*}

\begin{figure*}[h]
    \centering
    \includegraphics[width=0.75\linewidth]{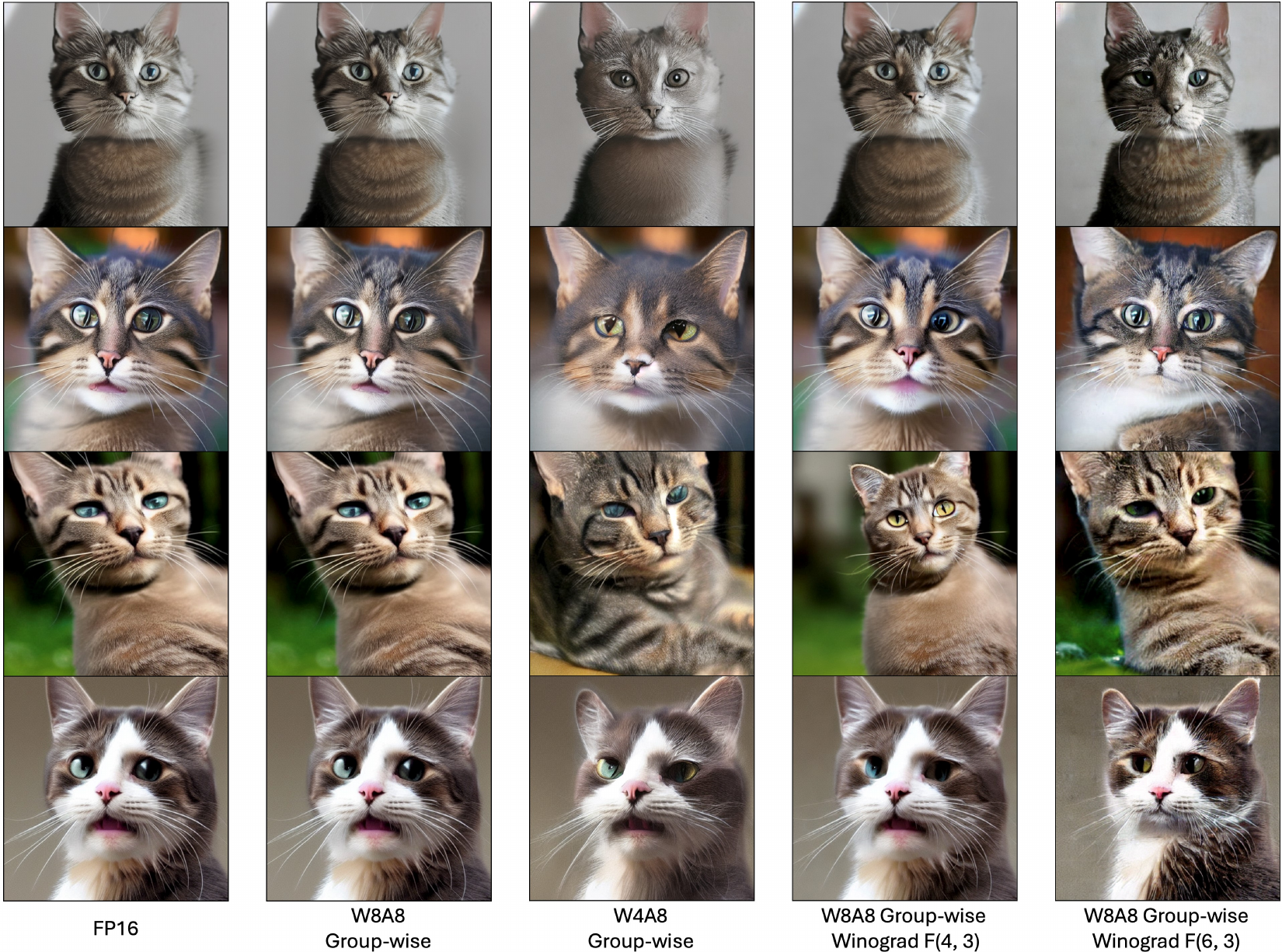}
    \caption{Stable Diffusion V1.5 with AKL and DPMSolver++ sampler. Prompt \textit{"A realistic photo of a lovely cat"}}
    \label{fig:cat-sdv1.5-comparison}
\end{figure*}

\begin{figure*}[t]
    \centering
    \includegraphics[width=0.9\linewidth]{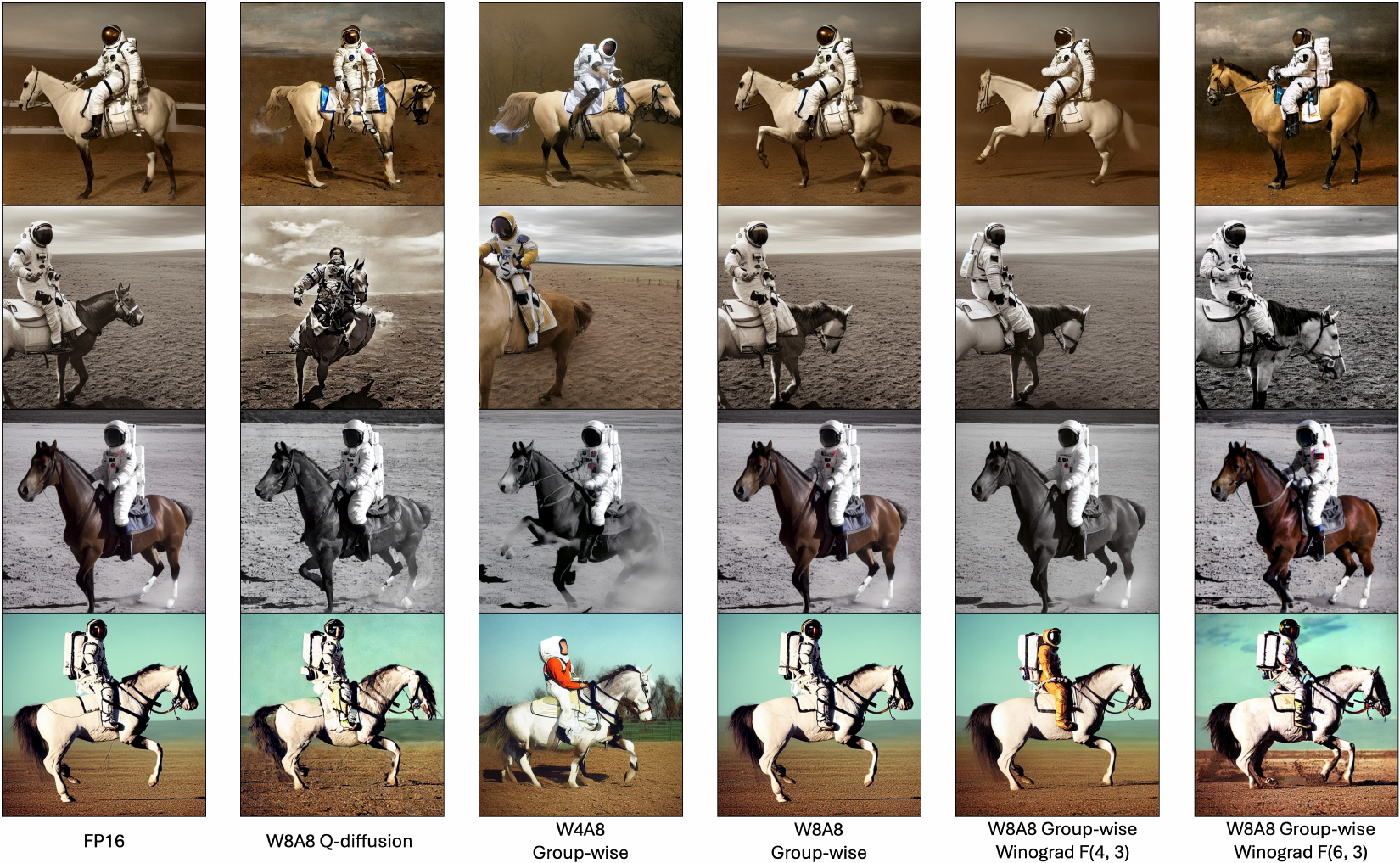}
    \caption{Stable Diffusion V1.4 with AKL and DPMSolver++ sampler. Prompt \textit{"A photograph of an astronaut riding a horse"}}
    \label{fig:astronaut-sdv1.4-comparison}
\end{figure*}

\begin{figure*}[h]
    \centering
    \includegraphics[width=0.9\linewidth]{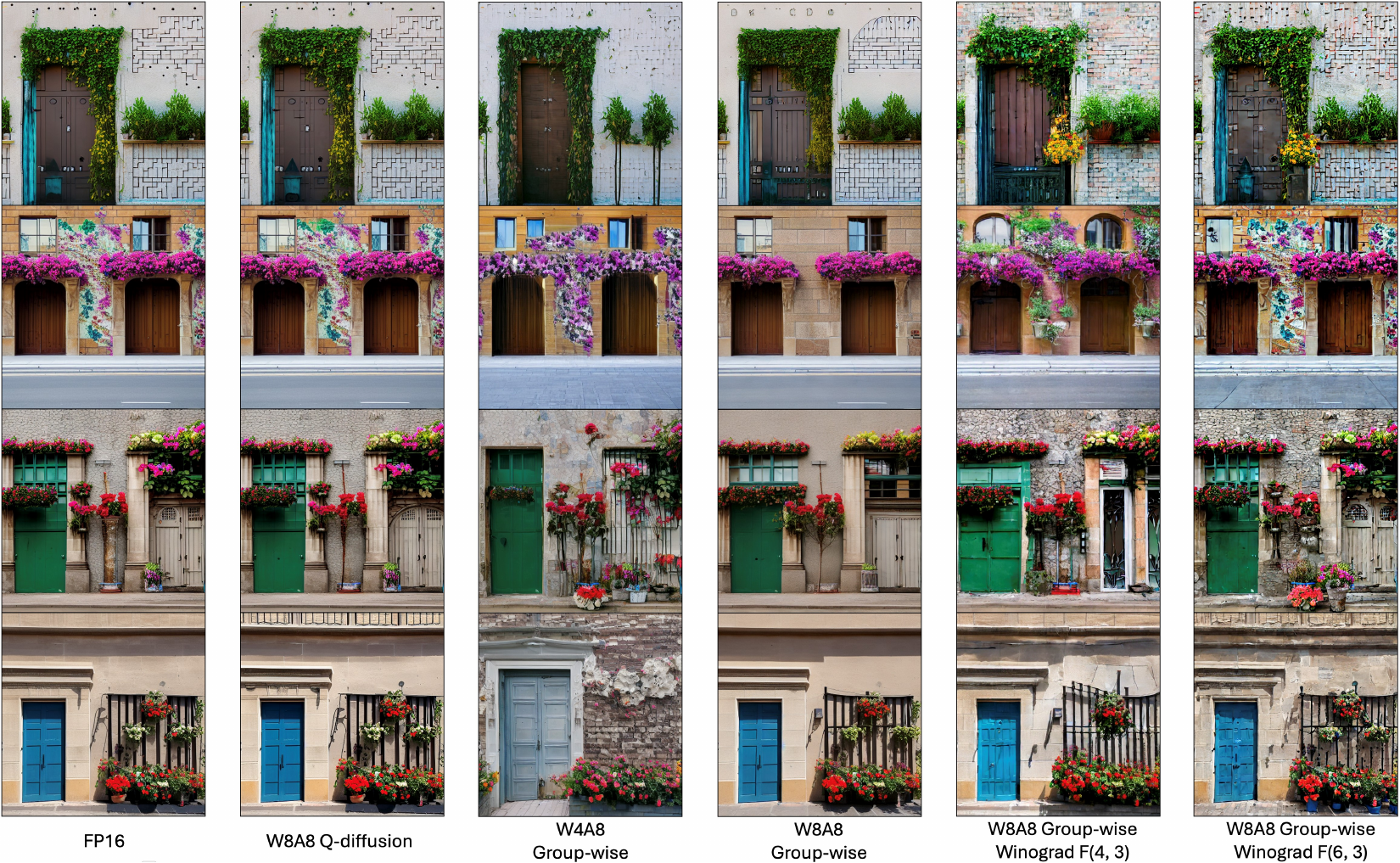}
    \caption{Stable Diffusion V1.4 with AKL and DPMSolver++ sampler. Comparison with Q-Diffusion. Prompt \textit{"A building wall and pair of doors, along with vases of flowers on the outside of the building"}}
    \label{fig:building-sdv1.4-comparison}
\end{figure*}

Figure~\ref{fig:dogs-instaflow-comparison}, \ref{fig:motor-instaflow-comparison}, \ref{fig:painting-sdv1.5-comparison}, \ref{fig:cat-sdv1.5-comparison}, \ref{fig:astronaut-sdv1.4-comparison}, and \ref{fig:building-sdv1.4-comparison} compare more qualitative examples generated by the $8$-bit Winograd convolution with those generated by the full-precision, 4-bit, and 8-bit models with standard convolution. We also show the images generated by $8$-bit Q-diffusion using standard convolution. It can be observed that W8A8 group-wise quantized Stable Diffusion can generate images of nearly identical quality compared with either the FP16 or Q-Diffusion model while being calibration-free. For W4A8, group-wise quantization shows no significant drop in image generation quality and text image alignment. 

\begin{table}[t]
    \caption{Train Winograd scales and transforms using calibration data.}
    \centering
    \begin{small}
    \begin{tabular}{ccccc}
        \toprule
        \multicolumn{5}{c}{\small IF-0.9B, COCO2017-5k, AKL} \\
        \hline
        Model & tile size & Bits & FID($\downarrow$) & CLIP($\uparrow$) \\
        \hline
        Learned scales & F(4,3) & 8/8 & 24.51 & 29.87 \\\cline{2-5}
        noise & F(6,3) & 8/8 & 26.58 & 29.65 \\
        \hline
        Learned scales & F(4,3) & 8/8 & 24.48 & 29.85 \\ \cline{2-5}
        Calibration data E2E & F(6,3) & 8/8 & 36.18 & 29.28 \\
        \hline
        Learned transforms & F(4,3) & 8/8 & 138.94 & 21.05 \\ \cline{2-5}
        Calibration data E2E & F(6,3) & 8/8 & 262.10 & 14.20 \\
        \toprule
        \multicolumn{5}{c}{\small SD-1.4, COCO2017-5k, AKL, DPMSolver++} \\
        \hline 
        FP16  & N/A & 16/16 & 21.63 & 31.72 \\
        \hline
        Learned scales noise & F(6,3) & 8/8 & 20.75 & 31.57 \\
        \hline 
        Learned scales BRECQ & F(6,3) & 8/8 & 170.24 & 23.24 \\
        \bottomrule
    \end{tabular}
    \end{small}
    \label{tab:train-e2e}
\end{table}

\begin{table} [h!]
    \caption{Results on the Stable Diffusion V1.4 model with group-wise quantization, AKL autoencoder and DPMSolver++ sampler with 25 steps. Comparison with the results from Q-Diffusion \cite{Li_2023_ICCV}. }
    \centering
    \begin{small}
    \begin{tabular}{lcccc}
    \toprule
        \multicolumn{5}{c}{SD-1.4, COCO2017-5k, AKL, DPMSolver++} \\
        \hline
        Model & tile size & Bits & FID($\downarrow$) & CLIP($\uparrow$) \\
        \hline
        FP16 & N/A & 16/16 & 21.63 & 31.72 \\
        \hline
        W4A8 & N/A & 4/8 & 21.21 & 30.84 \\
        \hline
        W8A8 & N/A & 8/8 & 21.52 & 31.70 \\
        \hline
        W8A8 & \multirow{2}{*}{N/A} & \multirow{2}{*}{8/8} & \multirow{2}{*}{21.03} & \multirow{2}{*}{31.16} \\ 
        Q-Diffusion \cite{Li_2023_ICCV} \\
        \bottomrule
    \end{tabular}
    \end{small}
    \label{tab:sdv14-akl-qdiffusion-dpm}
\end{table}

\begin{table} [h!]
    \caption{Results on the Stable Diffusion V1.4 model with group-wise quantization, AKL autoencoder and PLMS sampler with 25 steps. Comparison with the results from Q-Diffusion \cite{Li_2023_ICCV}. }
    \centering
    \begin{small}
    \begin{tabular}{lcccc}
    \toprule
        \multicolumn{5}{c}{SD-1.4, COCO2017-5k, AKL, PLMS} \\
        \hline
        Model & tile size & Bits & FID($\downarrow$) & CLIP($\uparrow$) \\
        \hline
        FP16 & N/A & 16/16 & 22.94 & 31.74 \\
        \hline
        W4A8 & N/A & 4/8 & 22.19 & 30.01 \\
        \hline
        W8A8 & N/A & 8/8 & 22.48 & 31.68 \\
        \hline
        W8A8 & \multirow{2}{*}{N/A} & \multirow{2}{*}{8/8} & \multirow{2}{*}{24.42} & \multirow{2}{*}{31.07} \\ 
        Q-Diffusion \cite{Li_2023_ICCV} \\
        \bottomrule
    \end{tabular}
    \end{small}
    \label{tab:sdv14-akl-qdiffusion-plms}
\end{table}

\section{Transferability of learned Winograd scales across datasets}
Because Winograd scales are learned from random noise inputs in our method, the same Winograd scale used for one dataset should be transferable to other datasets. 
Table~\ref{tab:resnets-results-cifar10} shows the accuracy results for the CIFAR10 dataset, whereas the same Winograd scale values learned for the ImageNet dataset models previously are applied to the CIFAR10 dataset models directly. Winograd, using learned scales, can successfully restore the accuracy of the full-precision ResNet networks for both tile sizes. This confirms the transferability of our learned Winograd scales across datasets.
Table \ref{tab:sd15-mjhq} shows the results from Stable Diffusion V1.5 with learned Winograd scales, using DPMSolver++ sampler with 25 sampling steps and a classifier-free guidance scale of 5.0 on the MJHQ dataset. 
The same learned Winograd scales work for both the COCO and MJHQ datasets.

\begin{table}[t]
    \caption{ResNets Top-1 accuracy results on the CIFAR-10 dataset.}
    \centering
    \begin{small}
    \begin{tabular}{cccccccc}
         \toprule
         \multirow{2}{*}{Model} & \multirow{2}{*}{tile size} & \multirow{2}{*}{Bits} & Standard & Winograd with \\
          & & & Winograd & Learned Scales \\
         \hline
         ResNet18 &  F(4,3) & W8/A8 & 92.55\% & 93.00\% \\
         93.07\% &   F(6,3) & W8/A8 & 33.62\% & 92.70\% \\
         \hline
         ResNet34 &  F(4,3) & W8/A8 & 93.05\% & 93.26\% \\
         93.33\% &  F(6,3) & W8/A8 & 22.90\% & 92.90\%\\
         \hline
         ResNet50 &  F(4,3) & W8/A8 & 93.49\% & 93.70\% \\
         93.65\% &  F(6,3) & W8/A8 & 81.13\% & 93.56\% \\
         \bottomrule
    \end{tabular}
    \end{small}
    \label{tab:resnets-results-cifar10}
\end{table}

\begin{table} [t]
    \caption{Results on the Stable Diffusion V1.5 model with group-wise quantization, AKL autoencoder and DPMSolver++ sampler with 25 steps.}
    \centering
    \begin{small}
    \begin{tabular}{ccccc}
    \toprule
    \multicolumn{5}{c}{SD-1.5, MJHQ-2k, AKL, DPMSolver++} \\
    \hline
    Model & tile size & Bits & FID($\downarrow$) & CLIP($\uparrow$) \\
    \hline
    FP16 & N/A & 16/16 & 41.40 & 25.78 \\
    \hline
    Learned scales & F(4,3) & 8/8 & 44.03 & 25.53 \\\cline{2-5}
    noise & F(6,3) & 8/8 & 45.56 & 25.56 \\
    \bottomrule
    \end{tabular}
    \end{small}
    \label{tab:sd15-mjhq}
\end{table}

\section{Comparison to prior QAT works for fully quantizing Winograd}

In comparison to previous QAT studies~\cite{fernandez2020searching} that achieve full quantization by learning Winograd transformation matrices, we can achieve comparable results by fine-tuning only the Winograd scales. As shown in Table~\ref{tab:resnets-results-cifar10}, $8$-bit ResNet18 with Winograd achieves an accuracy of $93\%$ for $F(4, 3)$ (an accuracy drop of $0.07\%$ in comparison to the full-precision model) using our learned scales, whereas Winograd-aware QAT~\cite{fernandez2020searching} observes an accuracy drop of $0.7\%$ for the similar setting.

\section{Highly optimized kernels design for text-to-image generation inference on CPUs}
While C/C++ runtimes like stablediffusion.cpp~\cite{sdcpp} demonstrate performance and potential on CPUs, the baseline group quantized kernels have significant compute overheads (the leftmost bar in Figure 6(a)). As a result, a higher proportion of compute instructions do not perform multiplies, i.e., real work, rendering them unsuitable for meeting the required latency requirements for text-to-image generation model variants deployed on commodity CPUs.

To reduce compute overhead and increase the reuse of the input matrices, as well as the use of MAC and vector operations, our highly optimized matrix multiply kernels consider a series of consecutive rows and columns from the two input matrices at once. The use of multiple rows and columns allows for greater reuse of quantized matrices and associated scale factors, as well as fewer load operations. Furthermore, to reduce the high overhead from reduction operations and quantized operand unpacking operations in group-quantized matrix multiply kernels, as well as improve MAC unit utilization and thus the percentage of useful work, our highly optimized kernels perform the following optimizations:
\begin{itemize}
\item Amortize the cost of loading operands and the cost of weight unpacking across multiple output channels, resulting in fewer load operations, increased operand reuse, and greater use of vector operations.
\item Eliminate the overhead of frequently occurring explicit reduction operations in the matrix multiply kernel (required to accumulate partial dot products from different vector lanes of SIMD fused multiply-accumulate (dot product) operations to obtain the final dot product result for a quantized weight group) of group-wise quantized kernels by ensuring that different vector lanes of dot product instructions operate on weight elements and groups from different output channels so that dot product instructions can take advantage of their implicit accumulate (reduction) operations.
\item Solve the resulting non-contiguous access patterns of quantized groups across channels in memory by reformatting weights offline to match the compute order and packing, then storing them in memory in reordered format.
\item Maximize the use of the available high MAC throughput matrix-matrix multiply-accumulate instruction, which can perform twice as many MAC operations when compared to an equivalent SIMD dot product instruction.
\end{itemize}

The benefits of our matrix multiply kernel optimizations for group-wise quantized Winograd convolution and attention layers for Arm CPUs will easily extend to GPUs, neural engines, and x86 processors for the following reasons: (1) Our kernels reduce the number of instructions in the critical path of the group-wise quantized kernels through efficient aligning and reordering of operands of group-quantized formats to take advantage of implicit reduction operations of vector dot product instructions, increase operand reuse, maximize MAC unit utilization, minimize overhead and memory accesses, (2) The vector and matrix-matrix multiply operations used in optimized group-wise quantized kernels are available in other existing CPU and GPU architectures, so there is no need to add new instructions to the CPU/GPU ISA.

Our code is available at \url{https://gitlab.arm.com/artificial-intelligence/efficientwinograd}.

